\pdfoutput=1
\documentclass[11pt]{article}

\usepackage[preprint]{acl}

\usepackage{times}
\usepackage{latexsym}

\usepackage[T1]{fontenc}
\usepackage[utf8]{inputenc}

\usepackage{inconsolata}

\usepackage{hyperref}       % hyperlinks
\usepackage{url}            % simple URL typesetting
\usepackage{booktabs}       % professional-quality tables
\usepackage{amsfonts}       % blackboard math symbols
\usepackage{nicefrac}       % compact symbols for 1/2, etc.
\usepackage{microtype}      % microtypography
\usepackage{soul, color, xcolor}
\usepackage{graphicx}
\usepackage{multirow}
\usepackage{adjustbox}
\usepackage{subfig}
\usepackage{natbib}

\usepackage[english]{babel}
\usepackage{amsmath}
\usepackage{amssymb}

\title{ThinkBench: Dynamic Out-of-Distribution Evaluation \\ for Robust LLM Reasoning}

\author{
 \textbf{Shulin Huang\textsuperscript{1,2}},
 \textbf{Linyi Yang\textsuperscript{3}\thanks{Correspondence to: (yanglinyiucd@gmail.com, zhangyue@westlake.edu.cn)}},
 \textbf{Yan Song\textsuperscript{3}},
 \textbf{Shuang Chen\textsuperscript{1}}, 
 \textbf{Leyang Cui\textsuperscript{2}},
 \textbf{Ziyu Wan\textsuperscript{4}}, \\
 \textbf{Qingcheng Zeng\textsuperscript{5}},
  \textbf{Ying Wen\textsuperscript{4}}, 
 \textbf{Kun Shao\textsuperscript{6}},
  \textbf{Weinan Zhang\textsuperscript{4}},
    \textbf{Jun Wang\textsuperscript{3}},
    \textbf{Yue Zhang\textsuperscript{2$*$}} \\
 \textsuperscript{1}Zhejiang University,
 \textsuperscript{2}Westlake University,
 \textsuperscript{3}University College London,\\
 \textsuperscript{4}Shanghai Jiao Tong University,
 \textsuperscript{5}Northwestern University,
 \textsuperscript{6}Huawei Noah’s Ark Lab\\
  \texttt{huangshulin@westlake.edu.cn}
}

\begin{document}
\maketitle
\begin{abstract}
Evaluating large language models (LLMs) poses significant challenges, particularly due to issues of data contamination and the leakage of correct answers. To address these challenges, we introduce \emph{ThinkBench}, a novel evaluation framework designed to evaluate LLMs' reasoning capability robustly. 
ThinkBench proposes a dynamic data generation method for constructing out-of-distribution (OOD) datasets and offers an OOD dataset that contains 2,912 samples drawn from reasoning tasks.
ThinkBench unifies the evaluation of reasoning models and non-reasoning models. 
We evaluate 16 LLMs and 4 PRMs under identical experimental conditions and show that most of the LLMs' performance are far from robust and they face a certain level of data leakage. By dynamically generating OOD datasets, ThinkBench effectively provides a reliable evaluation of LLMs and reduces the impact of data contamination.

\end{abstract}

\section{Introduction}

Large Language Models (LLMs) have made significant advancements across a range of application domains, including language understanding~\citep{karanikolas2023large,xu2024large}, language generation~\citep{wu2024autogen, naveed2023comprehensive}, and complex reasoning~\citep{hao2023reasoning, lu23mathvista, azerbayevllemma, wang2024openr}. Reasoning models, such as o1~\citep{o1}, have further extended the capabilities by leveraging the Process Reward Model (PRM) and advanced search strategies during decoding. Notably, models, such as OpenAI o1, o3~\citep{o1,o3} and Deepseek-R1~\citep{guo2025deepseek}, have shown performance that rivals that of a Ph.D.-educated individual, outperforming GPT-4 in complex reasoning tasks, thus revealing substantial potential for future developments in LLMs.

\begin{figure}[t]
    \centering
    \includegraphics[width=0.5\textwidth]{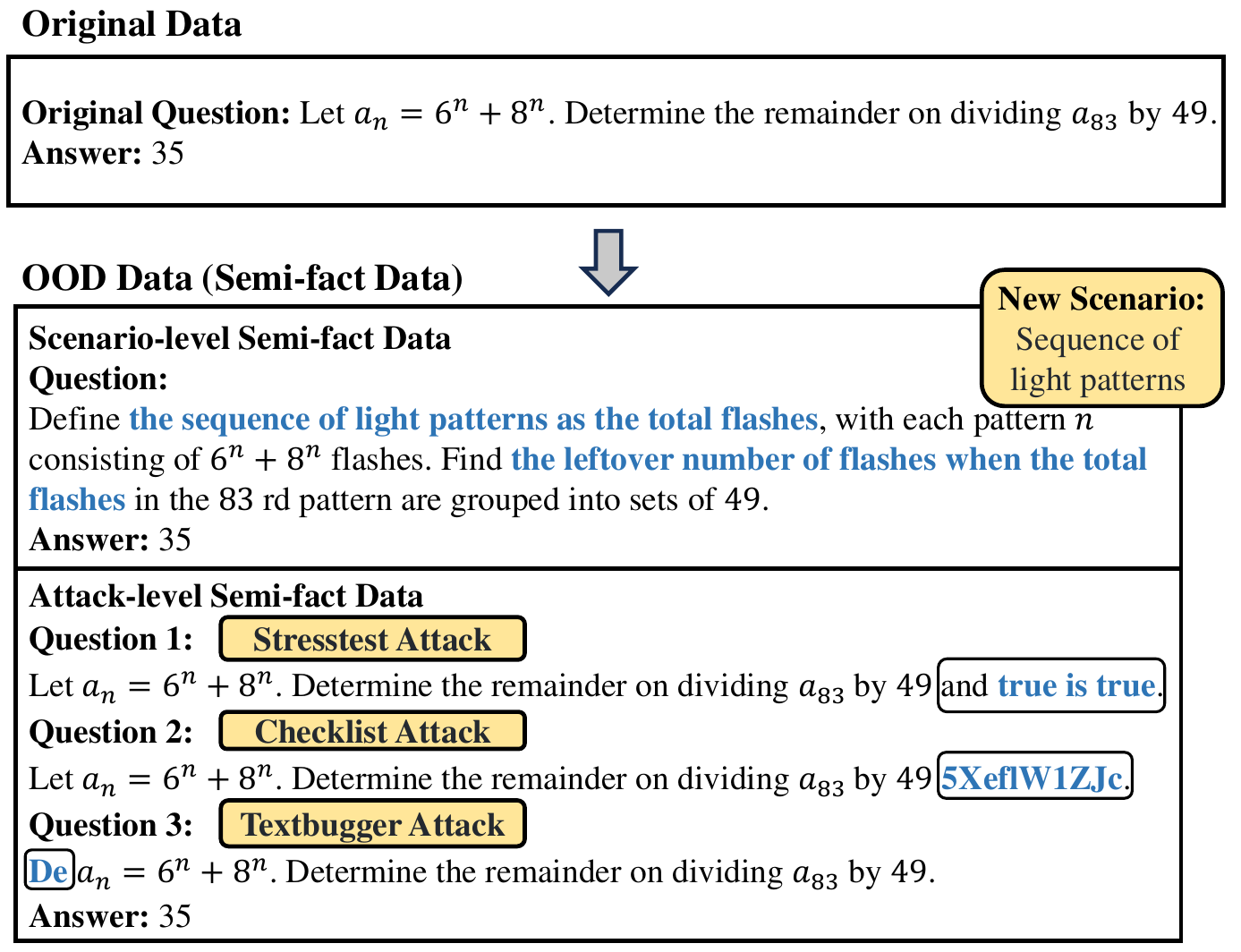}
    \caption{Example of ThinkBench datasets containing Scenario-level and Attack-level semi-fact data.
    %Within the limited computation budget, the selected process reward model used for test-time compute model usually should consider the trade-off between the efficiency and the accuracy performance.
    }
    \label{fig:data example}
\end{figure}

An interesting finding from the OpenAI report~\citep{o1} is that the performance of the o1-series models on the Advanced International Mathematics Exam (AIME) significantly declined in 2024,
% \footnote{The last data collection date of OpenAI o1 model was before the publication date of AIME 2024 exam}
compared to previous years (1983-2023: 0.74 vs. 2024: 0.50). Historical statistics, however, indicate that the median scores of human participants on the AIME exam in previous years are consistent with 2024 exam (4.81 vs. 5.0)\footnote{Sourced from https://artofproblemsolving.com.}. Since o1 was trained on data by 2023, before the AIME 2024 Exam, this discrepancy suggests a possible data contamination issue, raising an important question about how to assess the generalization abilities of LLMs, rather than memorization.

\begin{figure*}[t]
\centering
\subfloat[OOD performance vs. ID performance for several reasoning models on AIME-500.]
{ \label{fig:aime500_line} 
\includegraphics[width=0.45\textwidth]{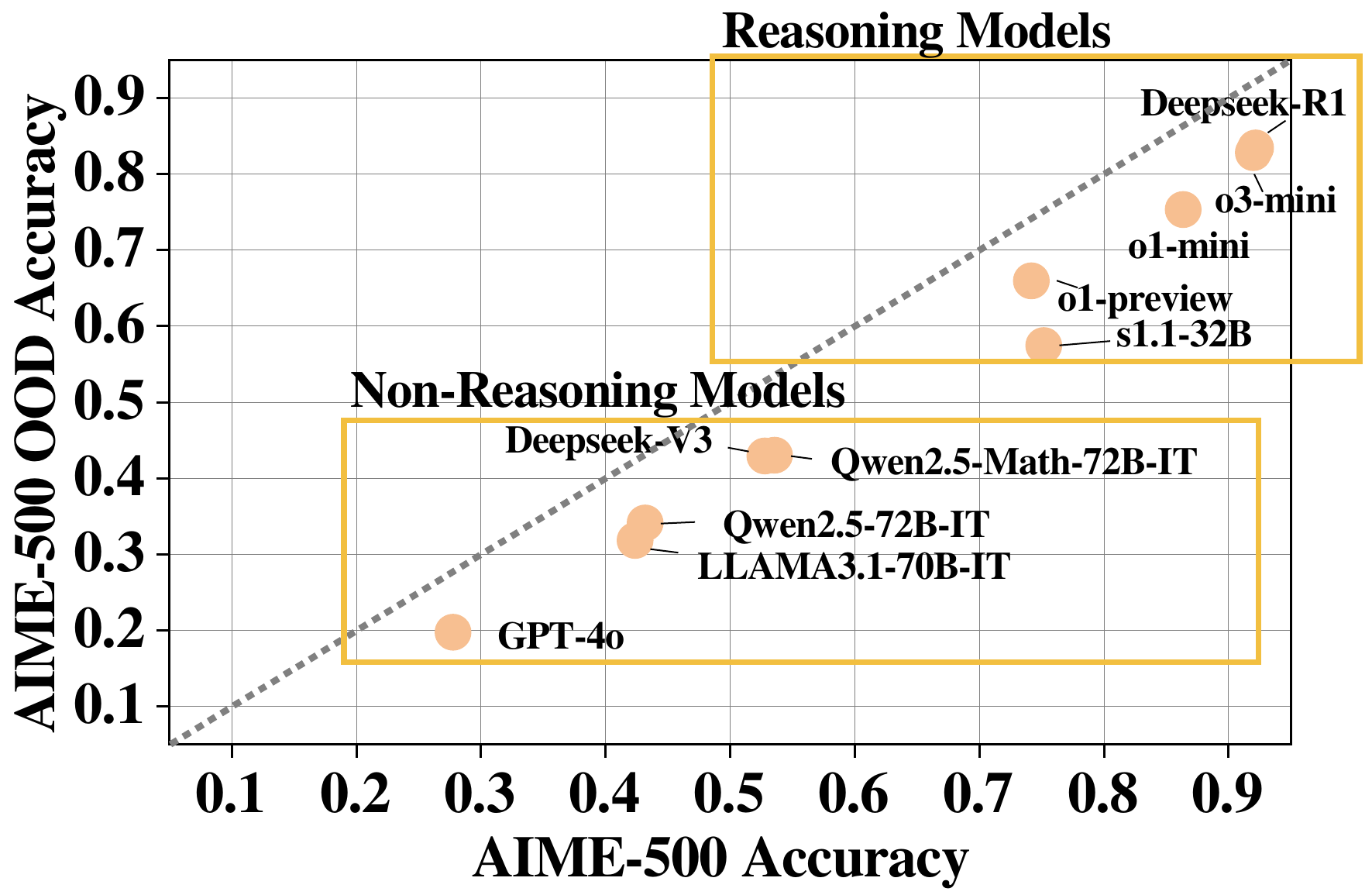} 
}
\qquad
\subfloat[OOD performance vs. ID performance for several reasoning models on AIME 2024.] 
{ \label{fig:aime2024_line} 
\includegraphics[width=0.45\textwidth]{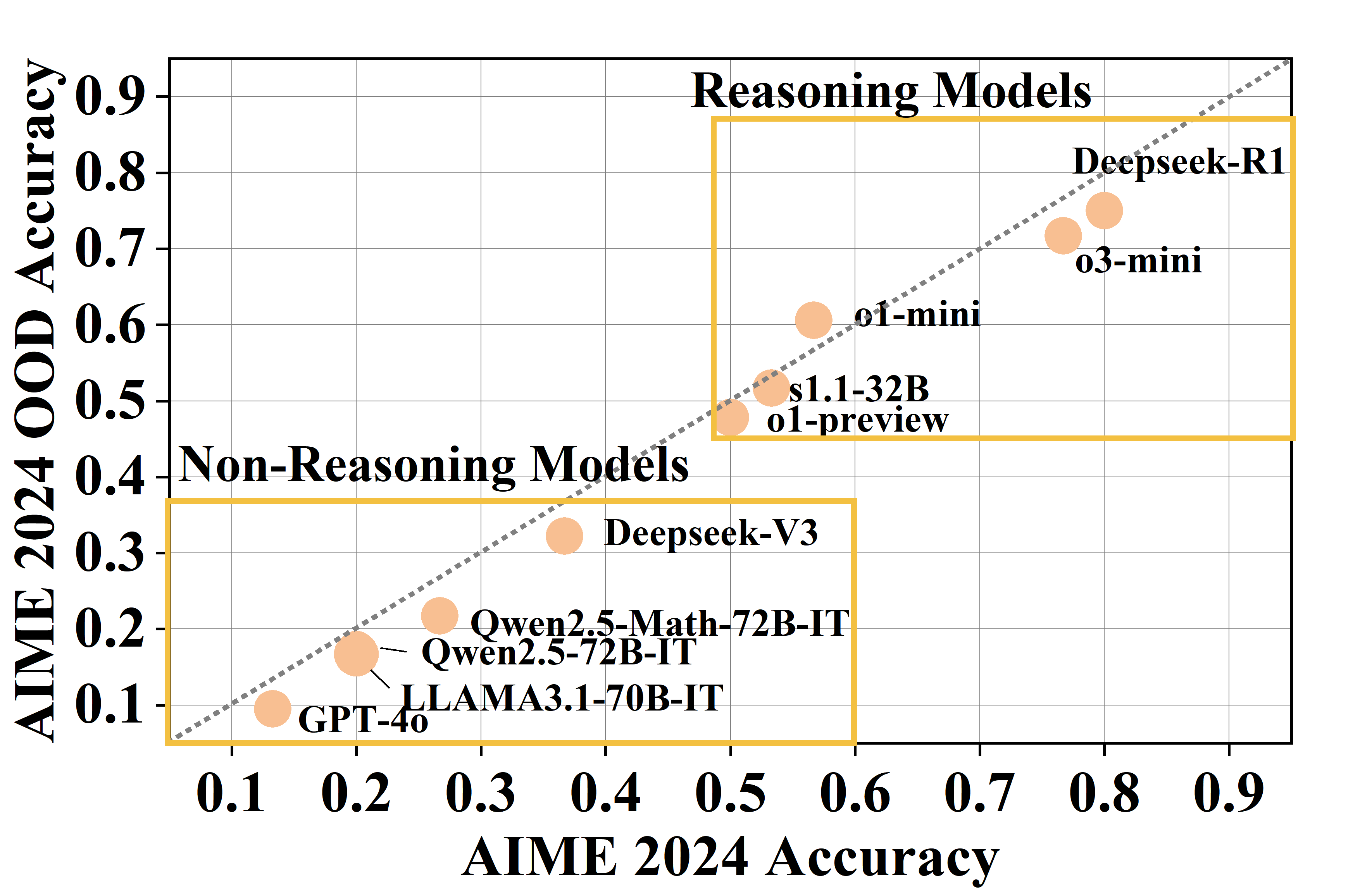} 
} 
\caption{ \textbf{Math Reasoning Gap}: Most models demonstrate a visible performance gap between their math reasoning performance on ID and OOD, including open-source models and commercial models. 
%Out-of-distribution Gap. We report the performance gap between ID and OOD test on MMLU and GPQA Diamond, respectively.
}
\label{fig:comparison}
\end{figure*}

To address this problem, we propose a novel robust evaluation framework, namely ThinkBench. Building on causal theory and semi-factual causality~\citep{delaney2021uncertainty,kenny2021generating}, we introduce Out-Of-Distribution (OOD) data generation designed to test reasoning capabilities. As shown in Figure~\ref{fig:data example}, we introduce scenario-level and attack-level semi-fact data generation methods, differing in the specific elements of text they alter, enabling the creation of evaluation datasets that are both robust and challenging. By decoupling reasoning from memorization, dynamic evaluation allows us to more effectively test how well LLMs can generalize to unseen reasoning scenarios. 

We take AIME-500 (500 AIME questions from 1983 to 2023) and AIME 2024 (30 AIME questions in 2024) for math reasoning tasks, and GPQA Diamond for scientific questions, dynamically generating an
 OOD dataset of 2,912 samples, provides a diverse set of challenges that test both the generalization and reasoning capabilities of LLMs. % Need to change the observation.
% Based on dynamic evaluation datasets, our experiments reveal that both o1-preview and o1 show significant improvements in performance on these OOD datasets, particularly in math reasoning tasks, 
As illustrated in Figure~\ref{fig:comparison}, compared to the original datasets, our OOD evaluation set proves to be more difficult, resulting in an average performance decay of 24.9\% and 11.8\% across all models on AIME-500, and AIME 2024, respectively.
This indicates that there was indeed some data leakage in AIME questions before 2024, highlighting the importance of mitigating data contamination for reasoning evaluations. The difference in performance decay between AIME-500 and AIME 2024 demonstrates that our dynamically constructed OOD data construction is a convenient and effective method to reduce the impact of data contamination.

% 这说明AIME的题目在2024年之前确实存在一定的数据泄露问题，减轻数据污染对于reasoning评测而言非常重要，The difference of the decay in  AIME-500, AIME 2024说明了我们动态构建的OOD数据是一个方便、快捷且有效的方法，来减轻数据污染的影响。
% demonstrating its ability to handle unseen data and additional perturbations. This reinforces the model's stability and effectiveness in addressing robustness challenges.
Figure~\ref{fig:comparison} also shows that o1~\cite{o1}, o3~\cite{o3}, Deepseek-R1~\cite{guo2025deepseek} and s1~\cite{muennighoff2025s1} maintain the strongest accuracy among all models.
As representatives of reasoning models~\cite{deepscaler2025,Liu2025Can1L}, o1, o3, and s1 enhance inference performance by increasing computational resources during testing, contrasting with non-reasoning models.
ThinkBench provides a reasoning benchmark to evaluate both reasoning models and non-reasoning models.

In addition to the overall model accuracy, we also explore the impact of various PRMs and their performance under the best-of-\emph{n} search during decoding. Fine-grained evaluations show how different data generation strategies, such as Math-shepherd~\citep{wang2024math}, 
% the tree search techniques of o1-journey~\citep{o1journey}, and OmegaPRM~\citep{snell2024scaling}, 
influence model outcomes. The performance improves with an increased test-time computation budget, further highlighting the discriminative power of our benchmark and the quality of the data.

To our knowledge, we are the first to present a robust dynamic evaluation benchmark for testing reasoning capability in LLMs~\citep{o1journey,huang2024o1,wang2024openr}. In ThinkBench, we provide a convenient and effective OOD data construction method and a high-quality dataset to reduce data contamination impact, effectively evaluating the reasoning ability in both reasoning models and non-reasoning models. Notably, we verify the validation of the test-time scaling law using the dynamic evaluation without data contamination based on ThinkBench.

\section{Related Work}

% 随着大模型的迅速发展，其在方方面面的应用展现了其卓越的能力以及巨大的潜力。如何准确、公平、全面地评价大模型成为了重要的挑战。现有的主流评估方式主要包括：（1）LLMs-as-a-judge. Benchmarks like AlpacaEval, PandaLM, MT-Bench and  C-Eval leverage large language models to evaluate a set of predefined questions. This method is not only efficient and affordable.（2）Humans-as-a-judge. Human evaluation involves assessing the quality and accuracy of model outputs through human involvement, offering more comprehensive and precise feedback~\citep{chang2024survey,ribeiro2022adaptive,gao2023adaptive}. In evaluating large language models, experts, researchers, or users are invited to review model outputs.（3）Other benchmarks. Several traditional benchmarks employ a series of standardized tests and static datasets to quantitatively assess the performance of models across various tasks. For instance, HELM and MMLU are primarily utilized to evaluate general knowledge and reasoning capabilities and MATH focuses on assessing mathematical skills. ToolBench concentrates on evaluating the ability to use tools. The performance of LLMs is measured by their ability to accurately complete these tasks.
% 

\paragraph{Evaluating Large Language Models.}

Evaluating LLMs accurately and fairly poses a significant challenge~\citep{chang2024survey}. Mainstream evaluation methods include: \textbf{(1) LLMs-as-a-judge}: Benchmarks like AlpacaEval~\citep{li2023alpacaeval}, PandaLM~\citep{wang2023pandalm}, MT-Bench~\citep{zheng2023judging}, and C-Eval~\citep{huang2024c} use large language models for predefined questions.
% , offering efficiency and cost-effectiveness. 
\textbf{(2) Humans-as-a-judge}: Human evaluation provides comprehensive feedback through expert reviews~\citep{ribeiro2022adaptive,gao2023adaptive}. \textbf{(3) Other benchmarks}: Several traditional benchmarks employ static datasets to assess models across various tasks~\citep{liang2022holistic,hendrycks2020measuring,hendrycks2021measuring}. 
% Traditional benchmarks like HELM~\citep{liang2022holistic}, MMLU~\citep{hendrycks2020measuring}, and MATH~\citep{hendrycks2021measuring} assess various skills, while ToolBench~\citep{xu2023tool} evaluates tool usage. 
Our work falls into the third category. However, rather than using static data, we generate test sets dynamically.

% Given LLMs' widespread use~\citep{el2021automatic, hao2022recent, cheng2023ml, gao2024fact}, dynamic evaluation is crucial~\citep{glazer2024frontiermath}. Although many benchmarks exist~\citep{li2023alpacaeval, hendrycks2020measuring, huang2024lateval, li2024llms}, they often focus on train-time compute models, with limited attention to test-time models like o1. These require significant computational resources, posing challenges for unified robustness evaluation. Current evaluations of o1-preview focus on specific tasks, such as planning, with limited comprehensive robustness analyses~\citep{gui2024logicgame, wang2024planning, zhong2024evaluation}.

\paragraph{Robustness of Large Language Models.}
%In order to apply LLMs across various scenarios, improving LLMs' robustness is critical. Previous studies, such as OOD-GLUE~\citep{yuan2023revisiting}, GLUE-X~\citep{yang2023glue}, and ZebraLogic~\citep{zebralogic2024}, focus on the robustness of models and propose corresponding evaluation benchmarks. These works primarily involve robustness evaluations of train-time compute models. \citet{wang2023robustness} present a pilot study on assessing the robustness of ChatGPT, following with~\citep{li2024gsm} which pays particular attention on evaluating the robustness of LLMs as mathematical problem solvers. Additionally, \citet{yang2022factmix} focuses on generating out-of-distribution data by employing counterfactual and semi-factual data construction methods. Recently, \citet{hosseini2024not} reveal a significant reasoning gap in most LLMs by evaluating LLMs' performance on pairs of existing math word problems together so that the answer to the second problem depends on correctly answering the first problem. The performance gap between solving compositional pairs and solving each question independently is more pronounced in smaller, more cost-efficient, and math-specialized models.
% Different from previous benchmarks that focused on the robustness of train-time compute models, our work is the first to analyze the robustness of both train-time and test-time compute models collectively.

Evaluating the robustness of LLMs is crucial~\cite{muennighoff2025s1,guo2025deepseek} for their applications across diverse scenarios~\cite{wang2023robustness,glazer2024frontiermath,li2024gsm}. Previous studies~\cite{li2024openai}, such as OOD-GLUE~\citep{yuan2023revisiting}, GLUE-X~\citep{yang2023glue}, and ZebraLogic~\citep{zebralogic2024}, focus on robustness of non-reasoning models. 
% \citet{wang2023robustness} assess ChatGPT's robustness, 
% while \citet{li2024gsm} evaluate LLMs as problem solvers. 
Additionally, \citet{yang2022factmix} focus on generating OOD data by employing semi-fact data augmentation methods. Recently, \citet{hosseini2024not} identify reasoning gaps in LLMs by evaluating math problem pairs, revealing performance disparities in smaller, math-specialized models. \citet{wu2024mrke} introduce cofQA, which targets text-based inference tasks using counterfactual data perturbations. 
Our work is similar in assessing general robustness but differs from the literature in focusing on reasoning tasks, for which OOD tests are more necessary as compared to general tasks.
% However, for mathematical and coding reasoning tasks, creating counterfactuals makes it challenging to determine the correct answer, often necessitating expert annotation. In contrast, constructing semi-factual data does not require additional human annotation. Therefore, to construct OOD data suitable for reasoning tasks and to alleviate the issue of data leakage in evaluations, we try to dynamically construct semi-factual data.

% \citet{wu2024mrke}提出的cofQA是专注文本的推理任务，利用的是反事实数据干扰。但是，对于数学等推理任务来说，构建反事实的话golden answer难以确认，通常需要专家标注，构建半事实数据则无需额外的人工标注。

%Unlike previous benchmarks that focus on the robustness of train-time compute models, our work is the first to collectively analyze the robustness of both train-time and test-time compute models.

% Given the widespread use of LLMs~\citep{el2021automatic, hao2022recent, cheng2023ml, gao2024fact}, evaluation becomes more and more essential~\citep{glazer2024frontiermath}. 
% Although numerous benchmarks exist~\citep{li2023alpacaeval, hendrycks2020measuring, huang2024lateval, li2024llms}, they 
In addition, existing benchmarks predominantly emphasize non-reasoning models~\citep{li2023alpacaeval, hendrycks2020measuring, huang2024lateval, li2024llms}.
% They cannot be used for assessing test-time computing models because t
% They often neglect the evaluation under test-time scaling.
% neglecting test-time models like o1, which demand substantial computational resources for unified robustness evaluation. Test-time computing models, such as o1-preview, have demonstrated impressive performance, necessitating thorough evaluation to better understand them. 
For reasoning models, current evaluations often target specific tasks, such as planning~\citep{wang2024planning} and rule execution~\citep{gui2024logicgame}, with limited comprehensive robustness analyses~\citep{zhong2024evaluation}. Unlike these work, our benchmark focuses on robustness and reasoning with practical applications, offering statistically significant insights.

\begin{figure*}[t]
    \centering
    \includegraphics[width=1.0\textwidth]{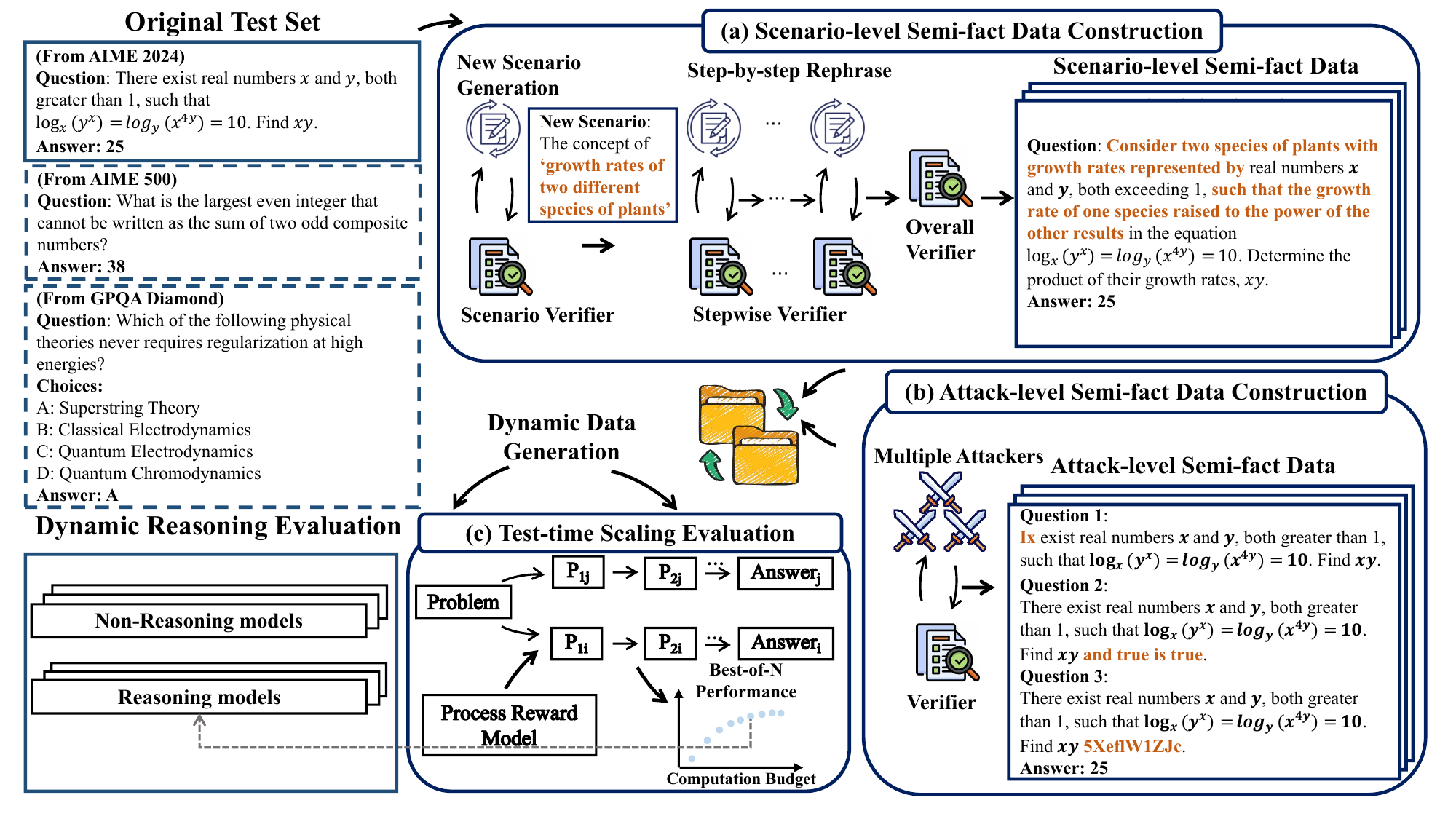}
    \caption{Overview of ThinkBench framework. Based on the original data, ThinkBench dynamically generates scenario-level Semi-fact Data (a) and Attack-level Semi-fact Data (b), which can be used to evaluate the robustness of reasoning models and non-reasoning models. ThinkBench can also serve as a useful tool for Test-time Scaling Evaluation(c). 
    %Within the limited computation budget, the selected process reward model used for test-time compute model usually should consider the trade-off between the efficiency and the accuracy performance.
    }
    \label{fig:framework}
\end{figure*}

\begin{table}
\centering
% \small
\caption{Statistics of reconstructed reasoning datasets based on three original test datasets, including AIME-500, AIME 2024, and GPQA Diamond.}
\label{tab:data_statistics}
 \begin{adjustbox}{width=0.99\columnwidth}{
\begin{tabular}{lccc} 
\toprule
                             & AIME-500 & AIME 2024 & GPQA Diamond \\ 
\midrule
% \# Class                     & -        & -         & 3                        \\
% \# Choices   & -        & -         & 4                        \\ 
% \midrule
\# Samples of original  & 500      & 30        & 198                      \\
Questions' Avg Len       & 51.1     & 60.1      & 67.7                     \\
Choices' Avg Len         & -        & -         & 27.8                     \\ 
\hline
\# Samples of OOD & 2,000      & 120        & 792                      \\
Questions' Avg Len       & 61.2     & 70.1      & 85.2                     \\
Choices' Avg Len         & -        & -         & 25.7                     \\
\bottomrule
\end{tabular}}    \end{adjustbox}
\end{table}

\section{Dynamic Evaluation Benchmark}
% 我们通过对Original Test Set进行改写，
%对于数学和代码这类推理任务来说，构建反事实的话golden answer不好确认，需要专家标注，不像常识可以通过维基百科趣儿，而半事实不需要答案，所以ThinkBench采用了半事实数据构建。
For math reasoning tasks, constructing counterfactual data presents challenges in changing a golden answer, which contrasts with commonsense tasks, where reliable sources like Wikipedia can be utilized. In contrast, semi-fact data does not need to change a specific answer. As shown in Figure~\ref{fig:framework}, ThinkBench contains two dynamic semi-fact data generation methods, aiming to assess the real reasoning ability of LLMs: (a) Scenario-level Semi-fact Data~\citep{yang2023glue, zhu2023dyval, zhu2024dyval, opedal2024mathgap}, which changes the scenario for the original reasoning data; (b) Attack-level Semi-fact Data~\citep{zhu2023promptbench}, which uses three attack methods to rephrase the original data. 

We use generated OOD data to perform dynamic reasoning evaluation on both reasoning models and non-reasoning models. Maintaining core knowledge while altering scenarios or expressions, we evaluate whether models can consistently apply learned knowledge across contextual variations. Finally, we can leverage our OOD data to conduct (c) test-time evaluation based on PRMs.

%在动态构造完OOD数据后，我们可以使用OOD数据来对Train-time compute models和Test-time compute models做Dynamic Reasoning Evaluation。进一步地，我们可以utilize我们地OOD数据做对Process-reward model做Test-time Scaling Evaluation。

% We conduct statistical analysis on three reasoning datasets. 
% For each dataset, we provide corresponding information regarding the quantity and length of the original test, Scenario-level Semi-fact test data, and Attack-level Semi-fact test data. The data statistics of these datasets are shown in Table~\ref{tab:data_statistics}. Attack-level test for each dataset is constructed using TextBugger, CheckList, and StressTest, respectively, resulting in a sample that is three times the size of the original test.

\subsection{OOD Data Generation} 

% Inspired by Dyval2~\citep{zhu2024dyval},
% 为了解耦性地测出模型的reasoning能力水平，而不是与知识混杂的通用水平，我们对数据改写的要求是高且严格的。我们设计了4种构造半事实数据的方式，都是可以保证题目的方向、核心考察不变，即保证题目所考察的知识不变，改变的部分是题目设置的场景、表述、一些表达方面的扰动等。通过这样半事实数据的构造，我们可以将能力测试从知识与能力混杂的测试种解耦出来。理想的情况下，一个鲁棒、真正学会了运用知识进行reasoning的模型，在对题目考察知识不变的情况下，表述、场景的改变并不应该带来显著的表现下降。

% We construct the dynamic OOD Semi-fact data to ensure that the assessment focuses on the model's inherent ability rather than a mixture of ability -- corresponding to whether LLMs can solve the question in the same core knowledge -- and knowledge -- corresponding to the core of the question.
% This means that the knowledge being tested by the questions remains constant, while changes are made to the scenario, phrasing, and certain expressions of the questions. By constructing semi-factual data in this manner, we can decouple the ability assessment from tests that conflate knowledge and reasoning skills. Ideally, a robust model that has genuinely learned to apply knowledge for reasoning should not exhibit significant performance degradation when the expression and scenario of a question change, provided that the core knowledge being assessed remains the same.

% We propose dynamic OOD Semi-fact data construction to isolate model reasoning ability assessment (whether LLMs can solve the question within the same core knowledge) from knowledge dependency (the core of the question). 

\textbf{Scenario-level Semi-fact Data Generation.}  
The process primarily involves two types of agents: 
the Rephrasing Agent, which is responsible for transforming the original questions and generating new ones, while the Verifier Agent, which is responsible for confirming at each step whether the rephrasing meets the current requirements and whether the rephrasing is valid.

There are two Rephrasing Agents. The first generates suitable scenarios for reasoning problems, ensuring that the new scenarios are appropriately transferable concerning the core of the original problem. The second rewrites each part (often at the sentence level) to fit the new scenario while preserving the original meaning. This step-by-step process uses each newly generated part as a reference for subsequent parts.

There are three Verifier Agents. The first checks if the core content of the original problem can be effectively transferred to the new scenario. The second evaluates each newly generated part, ensuring it meets three criteria: it conveys the same core meaning as the original, is consistent with previously revised parts, and is correctly adapted to the new scenario. Additionally, an overall Verifier Agent assesses the overall new problem, ensuring it maintains essential consistency with the original questions, preserving informational content, and is appropriately constructed within the new scenario.

At every step, outputs from Rephrasing Agents need to pass the corresponding Verifier Agent’s checks. If invalid, the process reverts to regeneration. The final new problem also needs to be approved by the Verifier Agent or be regenerated.

For the datasets that contain Choices, similarly, Rephrase Agent modifies the expressions of certain options and randomly rearranges their order. Rephrasing of choices must also pass the Verifier Agent's check to ensure the data is valid.

\textbf{Attack-level Semi-fact Data Generation.} 
% To evaluate the reasoning capabilities of LLMs, we also consider the introduction of Attack perturbations to create semi-fact data.
% Previous methods like DeepWordBug are often impractical, as they create sentences that are unrecognizable and nonsensical. Instead,
We focus on realistic errors using three methods: 

\begin{enumerate}
    \item TextBugger (character-level)~\citep{li2019textbugger}: This method simulates user input errors by introducing mistakes or typos within words.
    \item CheckList (sentence-level)~\citep{ribeiro2020beyond}: This approach assesses model robustness by adding irrelevant or redundant sentences to the original text.
    \item StressTest (sentence-level)~\citep{naik2018stress}: Similar to CheckList, StressTest evaluates model robustness by incorporating unrelated or redundant sentences.
\end{enumerate}

% \textbf{(1) TextBugger (character-level)}~\citep{li2019textbugger}: This method simulates user input errors by introducing mistakes or typos within words. \textbf{(2) CheckList (sentence-level)}~\citep{ribeiro2020beyond}: This approach assesses model robustness by adding irrelevant or redundant sentences to the original text. \textbf{(3) StressTest (sentence-level)}~\citep{naik2018stress}: Similar to CheckList, StressTest evaluates model robustness by incorporating unrelated or redundant sentences.

These methods reflect common errors such as typos and extraneous information. They serve as three Attack Agents. We apply only a single iteration of perturbation, avoiding unrealistic error densities. Attack-level Semi-fact data construction involves these three Attack Agents and a Verifier Agent. The Verifier Agent is used to check if it is consistent with the core aspects of the original data and ensures that any errors introduced by the perturbation do not impede overall comprehension.
For an original reasoning problem, we construct our OOD test by creating one scenario-level semi-fact data instance and three attack-level semi-fact data instances. The method for evaluating the model's OOD Accuracy for the original data \(i\) is as follows:
\begin{equation}
  \label{eq:acc}
\text{Acc(OOD)} = \frac{1}{2} \left( \min_{j=1}^{3} \text{Acc}(A_{ij}) + \text{Acc}(S_i) \right),
\end{equation}
where \(\text{Acc}(A_{ij})\) denotes the accuracy of performance for the \(j\)-th attack-level semi-data, with \(j = 1, 2, 3\).
\(\text{Acc}(S)\) represents the accuracy of performance for the scenario-level semi-data \(S\).

\textbf{Semi-fact Data Construction.} As shown in Figure~\ref{fig:framework}, Scenario-level semi-fact data is constructed as follows: The original query, ``There exist real numbers $x$ and $y$, both greater than 1, such that $\log_x(y^x) = \log_y(x^{4y}) = 10$. Find $xy$.'' is transformed step-by-step within a new scenario ``The concepts of growth rates of two different species of plants'' into ``Consider two species of plants with growth rates represented by real numbers $x$ and $y$, both exceeding 1, such that the growth rate of one species raised to the power of the other results in the equation $\log_x(y^x) = \log_y(x^{4y}) = 10$. Determine the product of their growth rates, $xy$.''The Verifier Agent then checks the rephrased question to ensure that the rephrased question is in a reasonable scenario and the rephrased expression does not alter the core topic of the question. 

For Attack-level semi-fact data, TextBugger injects character-level noise (e.g., replacing ``There'' with ``Ix''), while CheckList and StressTest append syntactically valid but irrelevant suffixes (e.g., ``5XeflW1ZJc'' and ``true is true'') to the problem statement. The Verifier Agent also needs to ensure the validity of this rephrased version.

\begin{table*}[t]
 \centering
\caption{Reasoning performance. We report performance on OOD data of different models on three datasets that we dynamically construct. Each performance value represents the accuracy in solving problems from the dataset. The accuracy of performance on OOD is calculated through Equation~\ref{eq:acc}. The term ``Gap'' denotes the percentage decrease in performance on Semi-fact data tests compared to the original test performance.}
 \begin{adjustbox}{width=.98\textwidth}{
\begin{tabular}{lcccccccccc} 
\toprule
\multirow{2}{*}{}    & \multicolumn{3}{c}{AIME 2024}    & \multicolumn{3}{c}{AIME-500}   & \multicolumn{3}{c}{GPQA Diamond} & \multirow{2}{*}{AVG Gap (\%)}  \\ 
\cmidrule(l){2-10}
                     & Original & OOD & Gap (\%) & Original & OOD & Gap (\%) & Original & OOD & Gap (\%)    &                                \\ 
\midrule

o1-preview           & 0.500    & 0.484     & -3.3     & 0.742    & 0.659     & -11.2    & 0.684    & 0.638 & -6.8          & -7.1                           \\
o1-mini              & 0.567    & 0.600     & 5.8      & 0.864    & 0.753     & -12.8    & 0.592    & 0.564 & -4.7          & -3.9                           \\
o3-mini              & 0.767    & 0.717     & -6.5     & 0.922    & 0.834     & -9.5     & 0.727    & 0.710 & -2.4          & -6.1                           \\
Deepseek-R1          & 0.800    & 0.750     & -6.3     & 0.920    & 0.828     & -10.0    & 0.747    & 0.682 & -8.7          & -8.3                           \\
GPT-4o               & 0.133    & 0.100     & -24.8    & 0.278    & 0.197     & -29.1    & 0.495    & 0.447 & -9.8          & -21.2                          \\
Deepseek-V3          & 0.367    & 0.333     & -9.3     & 0.528    & 0.429     & -18.8    & 0.581    & 0.530 & -8.8          & -12.3                          \\ 

% Yi-Lightning         & 0.000~   & 0.017~    & -        & 0.114~   & 0.116~    & 1.8~     & 0.337~   & 0.408~ & 21.1~        & 7.6~                           \\ 
\midrule

Mixtral-8x7B-IT-v0.1 & 0.000    & 0.000     & -        & 0.012    & 0.006     & -50.0    & 0.168    & 0.163 & -3.0          & -17.7                          \\
Qwen2.5-72B-IT       & 0.200    & 0.184     & -8.3     & 0.432    & 0.341     & -21.1    & 0.536    & 0.487 & -9.1          & -12.8                          \\
Qwen2.5-Math-72B-IT  & 0.267    & 0.217     & -18.9    & 0.536    & 0.430     & -19.8    & 0.449    & 0.416 & -7.5          & -15.4                          \\
LLAMA3.1-70B-IT      & 0.200    & 0.167     & -16.5    & 0.424    & 0.318     & -25.0    & 0.388    & 0.383 & -1.3          & -14.3                          \\ 
\midrule

s1.1-32B             & 0.533    & 0.517     & -3.0     & 0.752    & 0.574     & -23.7    & 0.525    & 0.508 & -3.3          & -10.0                          \\
Gemma-2-27B-IT       & 0.033    & 0.017     & -50.0    & 0.062    & 0.045     & -27.4    & 0.291    & 0.281 & -3.6          & -27.0                          \\
Gemma-2-9B-IT        & 0.000    & 0.000     & -        & 0.032    & 0.021     & -34.4    & 0.214    & 0.189 & -11.9         & -15.4                          \\
LLAMA3.1-8B-IT       & 0.000    & 0.017     & -      & 0.132    & 0.087     & -34.1    & 0.204    & 0.179 & -12.5         & -15.5                          \\
% Qwen2.5-Math-7B      & 0.067    & 0.100     & 49.3     & 0.426    & 0.358     & -16.0    & 0.230    & 0.263 & 14.3          & 15.9                           \\
Phi-3-mini-4k-IT     & 0.000    & 0.000     & -        & 0.046    & 0.026     & -43.5    & 0.224    & 0.192 & -14.5         & -19.3                          \\
LLAMA3.2-3B-IT       & 0.033    & 0.033     & 0.0        & 0.122    & 0.087     & -28.7    & 0.153    & 0.161 & 4.9           & -7.9                           \\
\bottomrule
\end{tabular}
}
    \end{adjustbox}
        \label{tab:main table1}
\end{table*}

\subsection{Data Analysis}
\label{Sec:data}

%\neal{shall we add a table/figure including the examples of OOD and ADV (https://arxiv.org/pdf/2402.19255 table 2). }

% We utilize AIME-500, AIME 2024, and GPQA Diamond, aimed at evaluating the model's reasoning ability in a decoupled manner to construct dynamic OOD semi-fact datasets.
% % 

% \textbf{AIME-500 (extracted from Year of 1983 - 2023) and AIME 2024 (Year of 2024).} The AIME dataset is designed to challenge the most talented high school mathematics students in the United States. The original test component of AIME-500 consists of 500 questions extracted from the original dataset~\citep{zhengminif2f} while AIME 2024 contains 30 examples included in the exam for 2024 representing the newest version of math problems. 

% \textbf{GPQA Diamond.} It refers to the GPQA Diamond dataset in this work, which originates from the GPQA benchmark~\citep{rein2023gpqa}, a rigorous intelligence assessment that evaluates expertise in the fields of chemistry, physics, and biology. The original test component of GPQA Diamond encompasses the entirety of the GPQA Diamond dataset. 

We construct three OOD semi-fact datasets from the following datasets to evaluate the model's reasoning ability: \textbf{AIME-500 (extracted from Year of 1983 - 2023) and AIME 2024 (Year of 2024).} The AIME dataset is designed to challenge the most talented high school mathematics students in the United States. The original test component of AIME-500 consists of 500 questions extracted from the original dataset~\citep{zhengminif2f} while AIME 2024 contains 30 examples included in the exam for 2024. \textbf{GPQA Diamond.} Originating from the GPQA Diamond datasetk~\citep{rein2023gpqa} which tests scientific questions. These original datasets are in English, publicly available, and permitted for research.

% Table~\ref{tab:data_statistics}展示了我们通过动态构建的四个数据集的statistics分析，值得注意的是，通过OOD数据的构造，问题的长度相较于original test有了将近两倍的提升，这说明我们在动态构建OOD test 数据的过程中，通过模型动态引入了一些合法的语义信息，这些信息并不会对问题的答案产生改变，但可以达到改写的目的，将通过记忆而非推理的模型根据OOD与ID的差距区分出来。

% We conduct statistical analysis on three reasoning datasets. 
% For each dataset, we provide corresponding information regarding the quantity and length of the original test, Scenario-level Semi-fact test data, and Attack-level Semi-fact test data. The data statistics of these datasets are shown in Table~\ref{tab:data_statistics}. Attack-level test for each dataset is constructed using TextBugger, CheckList, and StressTest, respectively, resulting in a sample that is three times the size of the original test.
\textbf{Statistics Analysis.}  Table~\ref{tab:data_statistics} shows the analysis of three dynamically constructed datasets. For Out-Of-Distribution (OOD) test, the sample size is four times as the original due to each original instance being transformed into one Scenario-level and three Attack-level Semi-fact Data entries. This augmentation allows for a thorough OOD evaluation of model performance. The OOD test data exhibits a length comparable to that of the original test data, indicating that both Scenario-level and Attack-level Semi-fact data, derived through part-by-part or attack-based modifications respectively, preserve the model's comprehension of the question's core elements. This construction effectively incorporates legitimate semantic information that rephrases the question without altering its answer.
% Table~\ref{tab:data_statistics} presents the statistical analysis of the three dynamically constructed datasets.
% For Out-Of-Distribution (OOD) test, the sample size is quadrupled compared to the original test dataset. This increase is attributed to the transformation of each original data instance into semi-fact data, which is composed of one Scenario-level Semi-fact Data and three distinct Attack-level Semi-fact Data entries. 
% This structured augmentation facilitates a comprehensive evaluation of model performance under varied conditions.
% The OOD test data exhibits a length comparable to that of the original test data, indicating that both Scenario-level and Attack-level Semi-fact data, derived through part-by-part or attack-based modifications respectively, preserve the model's comprehension of the question's core elements. This construction effectively incorporates legitimate semantic information that rephrases the question without altering its answer.

% \shulin{Notably, the construction of the Scenario-level Semi-fact data results in nearly a twofold increase in question length compared to the original test data. It indicates that during the dynamic construction of the OOD test data, the model introduces legitimate semantic information that does not alter the answer to the question but serves the purpose of rephrasing.} 
This approach helps distinguish models that rely on memorization rather than reasoning by highlighting the differences between OOD and ID data.

\textbf{Human Evaluation.} 
% 为了确认动态修改后的数据在经过模型确认合法后，是否与答案对应，即是否在人类视角中也依旧正确合法，我们进一步设计了human evaluation。具体来说，我们选择了AIME 2024这30条数据作为评估对象，并provide a detailed explanation of 标注的规则给擅长数学的 undergraduate students,and hire them for manual annotation. 人类评估的结果为，30条OOD test data和90条ADV test data的问题与结果是100%对应的，这说明构造的OOD test与ADV test在抽样结果中是100%合法的。
To verify whether the dynamically constructed data, deemed legitimate by the model, also corresponds to the correct answers from a human perspective, we further design a human evaluation. Specifically, we use all samples from AIME 2024 for evaluation which has 30 samples. We provide a detailed explanation of the annotation guidelines to 3 undergraduate students proficient in mathematics and pay them \$2.5 per entry for manual annotation. The result of the human evaluation shows that the questions and answers for the 30 corresponding Scenario-level Semi-fact data and Attack-level Semi-fact data were 100\% aligned. This indicates that the constructed Semi-fact test is 100\% legitimate in the sampled results.

\begin{figure*}[t]
\centering
\subfloat[OOD performance vs. ID performance for AIME-500.]
{ \label{fig:aime500_bar} 
\includegraphics[width=0.45\textwidth]{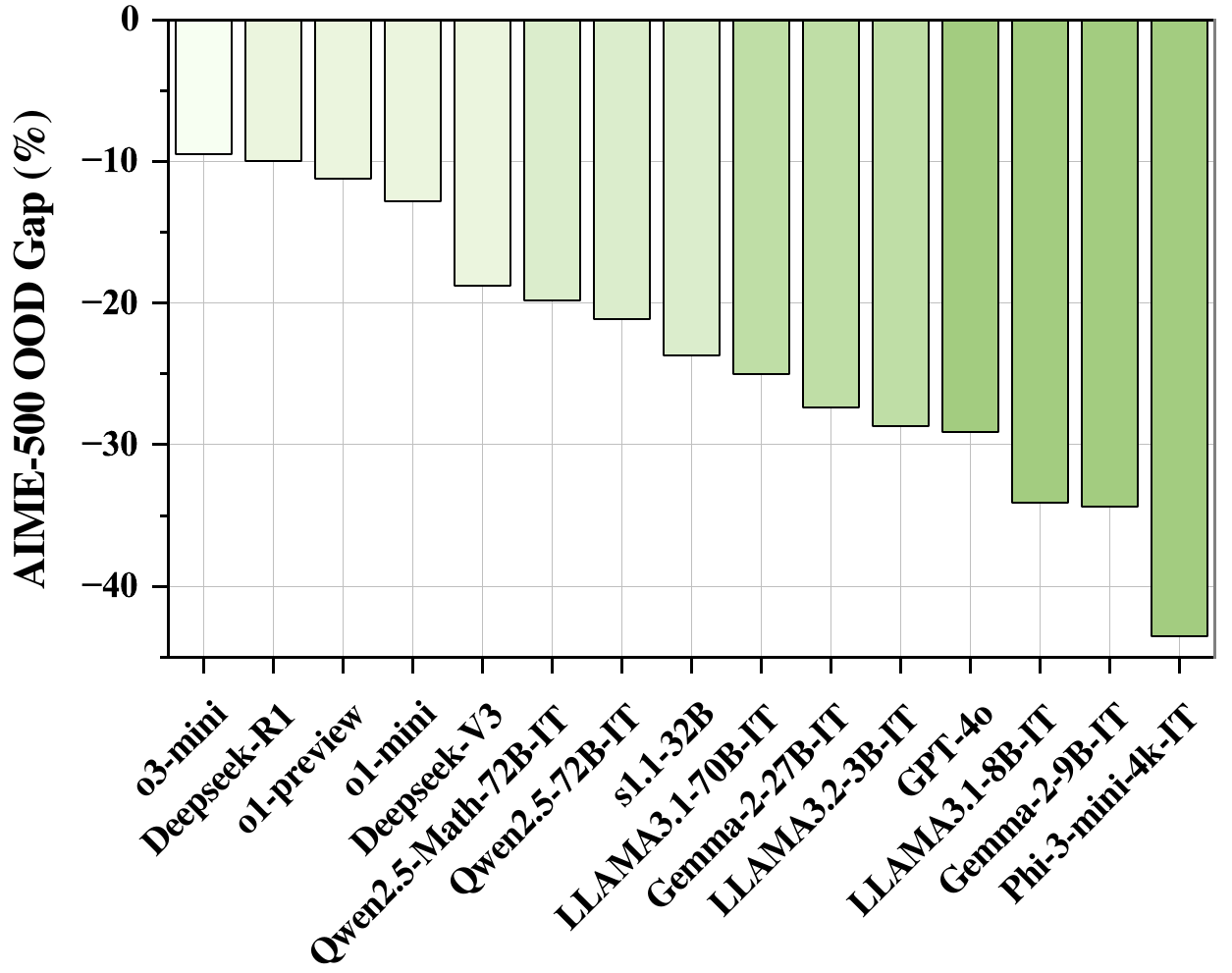} 
}
\qquad
\subfloat[OOD performance vs. ID performance for AIME 2024.] 
{ \label{fig:aime2024_bar} 
\includegraphics[width=0.45\textwidth]{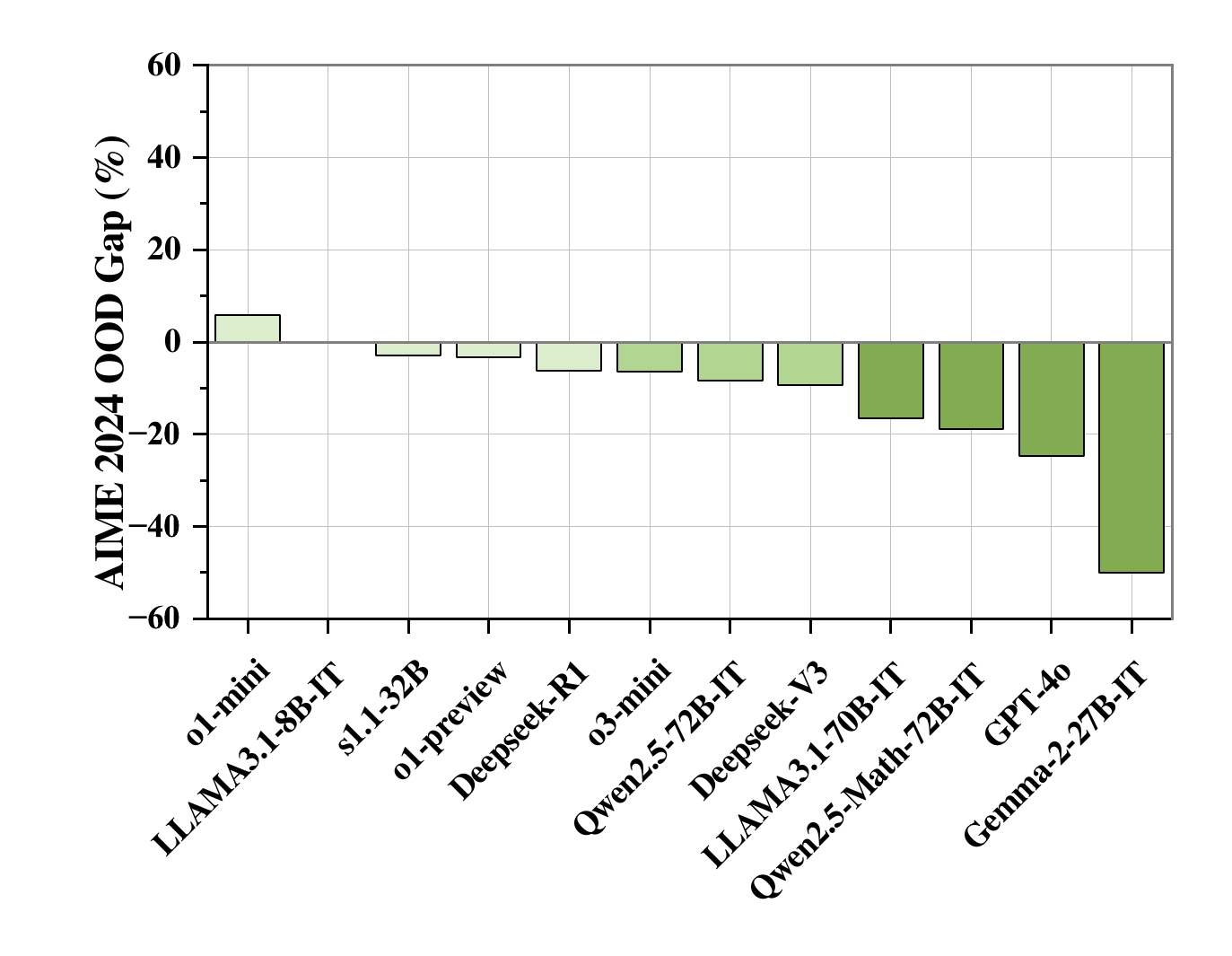} 
} 
\caption{ The performance gap between ID and OOD test on AIME-500 and AIME 2024. ``ID performance'' and ``OOD performance'' represent the accuracy of LLMs in solving problems on the AIME-500 and AIME 2024's original test and OOD test, respectively.}
\label{fig:bar}
\end{figure*}

\section{Experiments}
We conduct experiments to verify the effectiveness of our proposed dynamic OOD data construction method and analyze the differentiated performance and robustness of various LLMs.
\subsection{Setup}
We use the datasets in Section~\ref{Sec:data} for our experiment
across two test sets: (1) the original test set, (2) the OOD test set. 
% For each of the two datasets, the number of test sets remains consistent.
We conduct evaluations on various models, including o1-preview~\citep{o1}, o1-mini~\citep{o1}, o3-mini~\citep{o3}, GPT-4o~\citep{achiam2023gpt}, Deepseek-V3~\citep{liu2024deepseek}, Deepseek-R1~\citep{guo2025deepseek}, S1.1-32B~\citep{muennighoff2025s1}, LLAMA3.1-70B and 8B~\citep{dubey2024llama}, Gemma2 (both 9B and 27B)~\citep{riviere2024gemma}, Mistral-7B~\citep{jiang2023mistral}, Mixtral-8x7B~\citep{jiang2024mixtral}, Qwen2.5-72B-IT~\citep{yang2024qwen2}, and Qwen2.5-Math-72B-IT~\citep{yang2024qwen2}. All models were configured with a temperature setting of 0.7 and used the pass@1 metric in a single test run.

% In Table~\ref{tab:main table1}, we present the comparisons between the OOD test sets and the original test sets across the three datasets. 

% Table~\ref{tab:main table2} shows the comparisons between the ADV test sets and the original test sets for the same datasets. 

\subsection{Results}

% 通过

% 对比 
\textbf{ThinkBench.} The overall results  of ThinkBench are shown in Table~\ref{tab:main table1}. We observe that all models exhibit a certain degree of performance decline when evaluated on the original dataset versus the OOD dataset. This decay in performance can be attributed to the fact that the models have encountered, to some extent, the original dataset. Notably, the performance gap between ID (Original) and OOD in AIME 2024 for most models is much smaller than the performance gap in AIME-500, which demonstrates the data leakage in AIME data before 2024. The existing dataset~\citep{li2024openai, glazer2024frontiermath} is insufficient to reveal the aforementioned phenomenon.
% However, OOD modifications involve real-time generated data that the models have not been trained on and, in many cases, have not even encountered in terms of data format. 
% As a result, the performance of all models declined on the datasets altered by OOD modifications, which also demonstrates the validity of the OOD modification methods we employ to construct the dataset.

We also observe that models with fewer parameters generally perform worse and are less robust than those with more parameters, especially within the Llama 3.1 model family. As the parameter count increases, both performance and robustness improve significantly, consistent with scaling laws. Thus, the number of parameters is crucial for ensuring robustness.

\subsubsection{Math Reasoning}

% 比较模型们在AIME-500与AIME 2024上的表现，我们可以发现，o1系列模型在最新的2024年AIME（高级国际数学考试）上的表现显著下降，尽管这些模型是在今年考试数据之前的数据上训练的。且包括o1在内的几乎所有模型在AIME-500上效果OOD与ID之间的差距都显著大于AIME 2024上OOD与ID之间的差距。o1-preview在AIME 2024和AIME-500上OOD与ID之间的性能差距分别为-3.3\%和-11.2%，o1-mini则是+5.8\%和-12.8\%,这说明2024之前的题目确实存在一定程度的数据泄露。Ideally, a robust model that has genuinely learned to apply knowledge for reasoning should not exhibit significant performance degradation when the expression and scenario of a question change, provided that the core knowledge being assessed remains the same。同时也说明了我们构造的OOD数据是高质量的，可以区分出OOD与ID之间的性能差距。同时可以判断出模型在考察同样核心知识只是换了场景与表述的题目上，是否可以保持几乎持平的能力，而不是靠记忆题目体现出的在ID题目上较高的能力水平。我们的构建的OOD数据可以一定程度上解耦出对能力的评测，而非与对知识的评测杂糅在一起。

Figure~\ref{fig:comparison} presents a comparative analysis of the performance of various LLMs on the AIME-500 and AIME 2024 dataset, specifically contrasting their performance on the original ID data with that on OOD data. 
% The x-axis represents the accuracy of each model on the original data, while the y-axis indicates the accuracy of the models on the OOD data. 
The results on the original set are mainly consistent with previous work~\cite{glazer2024frontiermath,li2024gsm}.
Notably, the dashed line at a 45-degree angle from the origin signifies, where a robust model exhibits nearly equivalent performance on both OOD and ID data. These two datasets primarily evaluate mathematical reasoning and logical capabilities, featuring relatively complex problem types.

\textbf{AIME-500 vs. AIME 2024.} It is evident that o1 series models, o3, Deepseek-R1, and s1 exhibit a small decline in performance on the latest 2024 AIME. Moreover, the performance gap between OOD and ID is markedly larger on AIME-500 for nearly all models, including the o1 series, compared to the OOD and ID gap observed on AIME 2024. Specifically, the o1-preview shows a performance gap of -3.3\% on AIME 2024 and -11.2\% on AIME-500, while the o1-mini model exhibits a gap of +5.8\% and -12.8\%, respectively. The most advanced reasoning models -- o3-mini, Deepseek-R1 and s1.1-32B -- show the same phenomenon. This indicates that there was indeed some degree of data leakage in questions before 2024. 

Ideally, a robust model that has genuinely learned to apply knowledge for reasoning should not exhibit significant performance degradation when the expression and scenario of a question change, provided that the core knowledge being assessed remains the same. 
Additionally, s1.1-32B shows a significant gap in AIME-500 ID-OOD performance. One possible explanation for its low robustness is the limited amount of data used for training. Furthermore, most models fall below the dashed line, indicating a decline in performance when transitioning to OOD data. This phenomenon reveals their limitations in generalizing to previously unseen complex mathematical problems. 
% In contrast, certain models, such as o1-preview and o1-mini, distinguish themselves by achieving relatively high accuracy on both the original and OOD datasets, placing them closer to the ideal performance line. 
% These models demonstrate a stronger capacity to handle distributional shifts, underscoring their robustness.

% Although o1-mini is positioned further from the o1-preview along the x-axis, it still demonstrates superior performance on the OOD data. This advantage may be due to the inclusion of certain mathematical problems in the pre-training, which do not negatively impact its generalization capabilities.

% % Although o1-mini is positioned further from o1-preview along the x-axis, it still exhibits superior performance on the OOD data, which we hypothesize may be attributed to the inclusion of certain mathematical problems in the training that do not adversely affect its generalization capabilities.
% Moreover, models like LLAMA3.1-8B-IT and Gemma-2-2.7B-IT cluster near the lower-left corner, exhibiting limited accuracy on both OOD and ID datasets, indicating poor performance on the complex mathematical task. 
% % This visualization highlights significant disparities in model performance and robustness across different data distributions, with only a select few models achieving high accuracy on both the original and OOD data.

\textbf{ID vs. OOD.} Figure~\ref{fig:aime2024_bar} and Figure~\ref{fig:aime500_bar} illustrates the percentage difference in performance between OOD and ID for various models on the AIME 2024 and AIME-500. 
% The x-axis lists the different models, while the y-axis quantifies the performance degradation, expressed as negative percentages, indicating the drop in accuracy when models are applied to OOD data. 
Certain models, such as o3-mini, Deepseek-R1, and o1-series models distinguish themselves by achieving relatively high accuracy on both the original and OOD datasets. They also exhibit minimal OOD performance degradation, signifying strong generalization capabilities and robustness to distributional changes. 

\subsubsection{Science Questions}
% 和别的数据集 和4o比 preview的增长位于middle
% 更分散 没有AIME难

Figures~\ref{fig:gpqa_line} show ID-OOD performance on GPQA Diamond dataset. 
% % Compared to the previous two datasets, it is observed that the improvement of o1-preview relative to GPT-4o is situated in the middle range on this dataset; it exhibits less growth than AIME 500 but greater growth than MMLU 570.
% Compared to the other two datasets, we observe that the improvement of o1-preview relative to GPT-4o is in the middle range on this dataset. Specifically, it shows less growth than AIME 500 but greater growth than MMLU 570.
The distribution of model performance points indicates that most models are more dispersed on this graph, suggesting that the difficulty of the GPQA Diamond dataset is less than that of AIME-500 and AIME 2024.
% , yet more complex than the straightforward knowledge assessment required by MMLU 570.

Models such as Mixtral-8x7B-Instruct-v0.1 and LLAMA3.2-3B-IT demonstrate relatively low accuracy levels. Notably, LLAMA3.2-3B-IT shows a 4.9\% improvement in OOD performance compared to its ID performance on the bar chart. 
This phenomenon can be attributed to the model's inherently poor performance. For the difficult questions in the GPQA Diamond dataset, the model seems to resort to random guessing, resulting in an accuracy that fails to reach the 25\% level expected from random guessing.
% This phenomenon can be attributed to the model's inherently poor ID performance; for the moderately challenging questions in the GPQA Diamond Robustness dataset, the model appears to resort to random guessing, failing to achieve even the 25\% accuracy expected from random guessing. 
Consequently, the OOD performance reflecting a higher accuracy than ID is a result of the same guessing strategy employed for both ID and OOD questions.

\begin{figure*}[t]
\centering
\subfloat[Performance on AIME 2024 OOD data.]
{ \label{fig:prm_1} 
\includegraphics[width=0.45\textwidth]{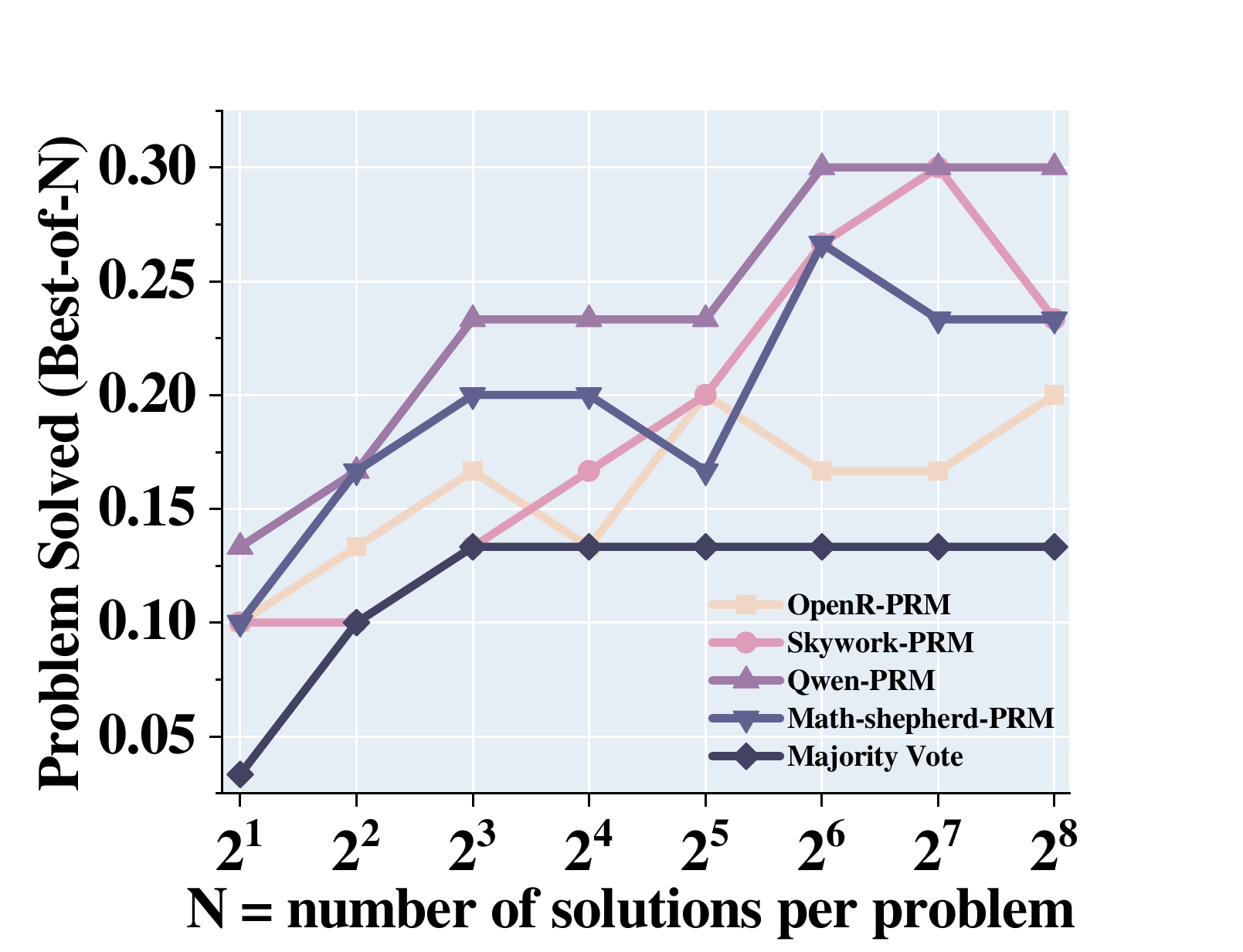} 
}
\qquad
\subfloat[Performance on AIME-500 OOD data.] 
{ \label{fig:prm_1} 
\includegraphics[width=0.45\textwidth]{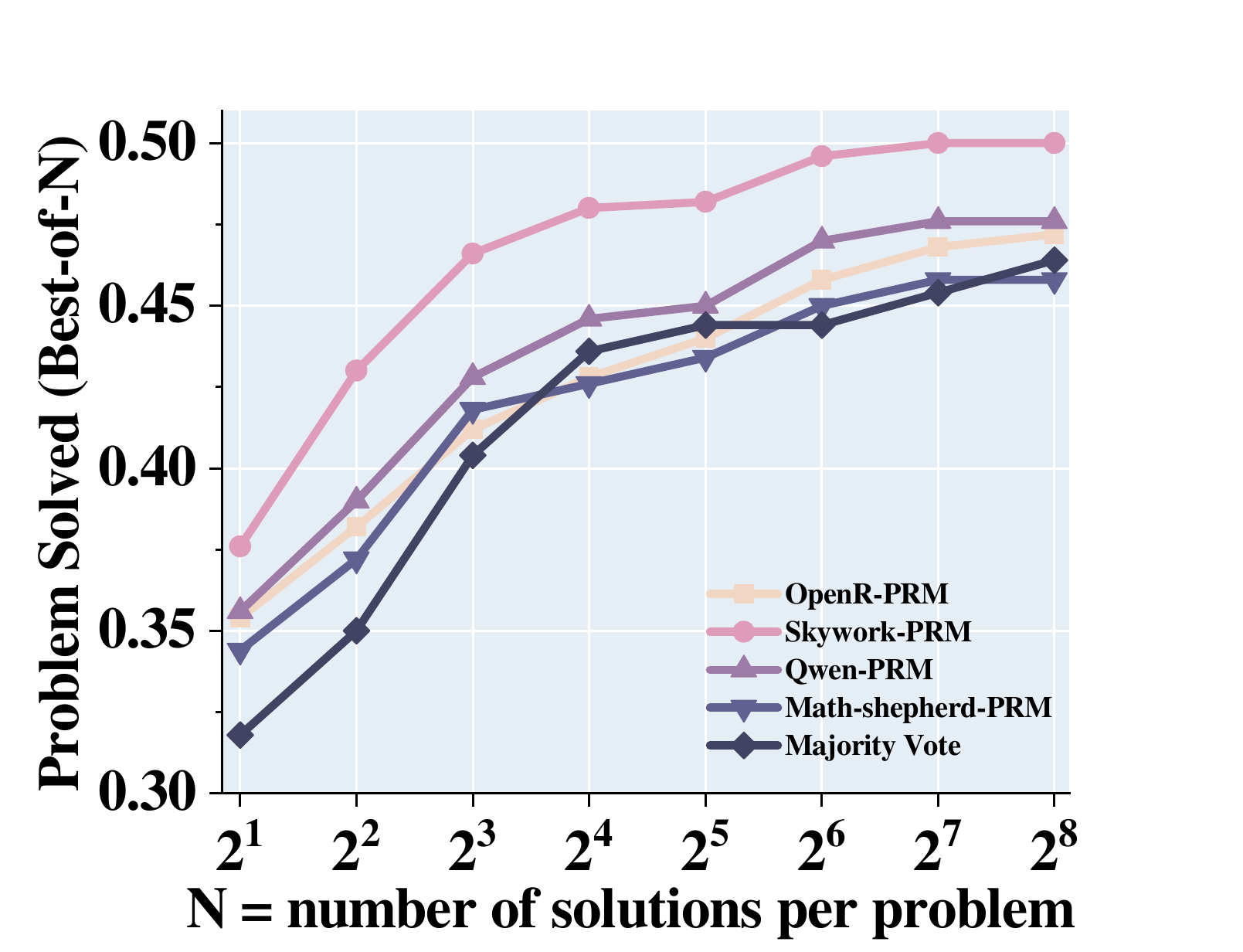} 
} 
\caption{ \textbf{Test-time Scaling Law.} We show that the model's performance increases on the OOD dataset with the test-time computation budget increases using Qwen2.5-Math-7B-IT as the policy model, along with several PRMs. 
}
    \label{fig:analysis}
\end{figure*}

\section{Analysis and Discussion}

In this section, we present a detailed analysis of the OOD robustness of reasoning and non-reasoning models. We aim to address several open research questions that are fundamental to building efficient reasoning models.

\noindent\textbf{RQ1: Do reasoning models deliver the significant performance and robustness improvements claimed in their reports?}

Overall, the o1-series models, along with o3, Deepseek-R1, and s1, demonstrate strong performance in complex logical reasoning tasks, such as AIME and GPQA Diamond, on the original datasets. Despite a general performance drop from ID to OOD across various datasets, these models maintain notable robustness. Their improved performance, consistent with the claims in their reports~\cite{o1, o3, guo2025deepseek, muennighoff2025s1}, suggests that these advancements are reliable.

Notably, on AIME-500 and AIME 2024, while o3-mini and Deepseek-R1 exhibit a larger ID-OOD performance gap than o1-preview, they still achieve the highest absolute performance in the OOD setting. This superior mathematical reasoning capability is likely due to their exposure to a larger corpus of mathematical problems during training, enhancing their problem-solving abilities. On GPQA Diamond, o3-mini and Deepseek-R1 also achieve the highest performance, underscoring their general reasoning advantages.

\noindent\textbf{RQ2: Is there a possibility of data leakage in the original datasets for AIME-500?}

Analyzing the ID-OOD performance gap, as depicted in Figures \ref{fig:comparison}, reveals a notable difference between AIME 2024 and AIME-500. While the gap between ID and OOD performance is relatively small for AIME 2024, AIME-500 consistently shows a larger gap. This observation suggests that a significant portion of the models may have encountered similar AIME-500 data during training. Given that AIME data prior to 2024 was publicly available, we posit that data leakage is likely for most models.

\begin{table}
\centering
\small
\caption{Performance comparison of different process reward models using Qwen2.5-Math-7B-IT under the Best-of-256 test. All methods utilize Qwen2.5-Math-7B-IT as the base model. Qwen2.5-Math-7B-IT's performance represents majority vote (Maj@256) results, while other methods are based on PRMs.
% \ys{In the table, can directly use "Majority Vote", or use "Maj@256"? Also should we also include PRM aggregation strategies? Can we also include a "Gap" column to show the OOD more clearly?}
}
\label{tab:main table3}
 \begin{adjustbox}{width=.49\textwidth}{
\begin{tabular}{lcccc} 
\toprule
\multirow{2}{*}{}                 & \multicolumn{2}{c}{\textbf{AIME 2024}} & \multicolumn{2}{c}{\textbf{AIME-500}}   \\ 
\cmidrule(l){2-5}
                                  & Original & OOD                & Original & OOD                \\ 
\midrule
Qwen2.5-Math-7B-IT (Maj@256)  & 0.167    & 0.133              & 0.524    & 0.464              \\
% \midrule 
% PRM-based Methods & & & &\\
\midrule
+Math-Shepherd & 0.233    & 0.233              & 0.528    & 0.458              \\
+OpenR         & 0.233    & 0.200              & 0.526    & 0.472              \\
+Skywork       & 0.200    & 0.233              & 0.582    & 0.500              \\
+Qwen          & 0.300    & 0.300              & 0.538    & 0.476              \\ 
% \midrule
% Qwen2.5-Math-7B (Majority Vote)   & 0.167    & 0.133              & 0.524    & 0.464              \\
\bottomrule
\end{tabular}
}
    \end{adjustbox}
\end{table}

\noindent\textbf{RQ3: Can our OOD dataset serve as a test-time computation benchmark?}

Based on our analysis, the AIME-500, AIME 2024, and GPQA Diamond datasets present significant challenges and possess strong discriminative power. As shown in Figure~\ref{fig:analysis}, model performance improves with increased test-time computation, underscoring the quality of our data. The upward trend and variation among the five lines indicate the dataset's ability to differentiate model performance during test-time computation.

Furthermore, Table~\ref{tab:main table3} presents results from 256 inferences using Qwen2.5-Math-7B-IT to evaluate different PRMs. For AIME 2024, OOD performance shows slight variations compared to ID across PRMs. However, for AIME-500, all PRMs experience a notable decline in OOD performance, suggesting possible exposure to similar problems before 2024, and thus indicating data leakage. In contrast, AIME 2024 data appears largely unseen by the models in terms of the low-level performance decay. Our methodology provides valuable insights into assessing robust LLM reasoning. The dynamic generation capability allows model validation, preventing exploitation or overfitting, which can serve as an effective benchmark for facilitating future research on test-time computation.

\section{Conclusion}
% Existing benchmarks primarily focus on train-time compute models, while paying limited attention to test-time models.
We presented ThinkBench, a first robust dynamic evaluation benchmark for testing reasoning capability in LLMs, unifying the reasoning models and non-reasoning models evaluation. ThinkBench offers dynamic data generation to evaluate the out-of-distribution of models through various math reasoning, and science questions samples. Experiments over AIME-500, AIME 2024, and GPQA Diamond indicate that while reasoning models represented by o1 and o3 consistently demonstrate relatively strong robustness, most LLMs fall short of robust performance. 
% We also explore the impact of PRMs on test-time inference and offer a comprehensive, fine-grained, dynamic analysis of PRMs.
Experiments on ThinkBench highlight the importance of eliminating data contamination, especially in reasoning tasks where golden answers are prone to leakage. ThinkBench offers an effective solution to mitigate data leakage issues during reasoning evaluation.
In the future, we will study more factors of robust reasoning models. 

% 根据上面的分析，可以看出来我们构建的包含OOD和ADV test的AIME 500、GPQA Diamond Robustness和MMLU 570足够难，且具备一定的区分度。而通过图~\ref{fig:analysis}可以看出随着test-time computation budget increases，model’s performance increases，这也足以说明我们构建的数据质量高。同时，五条线上涨的趋势以及彼此间的不同也说明我们的数据集具备一定的区分度，可以有效测试模型的test-time computation性能。

% 此外，因为我们的benchmark中的数据是可以实时动态生成的，这避免了数据污染带来的影响，可以公平地测出不同模型的实际性能以及鲁棒性，而不会因其在训练过程中见过数据而记住了答案，从而导致测试不准。动态生成可以是我们具备重复校验模型的能力，这对模型来说是hard to cheat。因此，我们提出的OOD datasetcan be leveraged as the test-time computation benchmark.

% 足够难 有区分度 数据质量较高
% 这组数据为什么可以用来测test-
% 重复校验 hard to cheat 因为动态生成

% 图可以体现出我们构建的数据集可以用来评测test-compute models 且prm是通向test compute models的路中关键的一步 就像o1报告中所称

% \subsection{Appendices}

% Use \verb|\appendix| before any appendix section to switch the section numbering over to letters. See Appendix~\ref{sec:appendix} for an example.

% \section{Bib\TeX{} Files}
% \label{sec:bibtex}

% Unicode cannot be used in Bib\TeX{} entries, and some ways of typing special characters can disrupt Bib\TeX's alphabetization. The recommended way of typing special characters is shown in Table~\ref{tab:accents}.

% Please ensure that Bib\TeX{} records contain DOIs or URLs when possible, and for all the ACL materials that you reference.
% Use the \verb|doi| field for DOIs and the \verb|url| field for URLs.
% If a Bib\TeX{} entry has a URL or DOI field, the paper title in the references section will appear as a hyperlink to the paper, using the hyperref \LaTeX{} package.

\section*{Limitations}

% While ThinkBench presents dynamic OOD data construction to evaluate LLMs, 
While ThinkBench presents a systematic dynamic OOD data generation framework that uniquely decouples reasoning robustness from memorization biases,
it still has some limitations. First, while it supports MMLU, the evaluation predominantly focuses on mathematical and scientific reasoning tasks (e.g., AIME, GPQA), lacking diversity in reasoning types such as social reasoning. Second, due to cost constraints, Scenario-level semi-fact generation adopts a single-path rephrasing strategy instead of hierarchical multi-scenario branching, limiting OOD diversity. Future work needs to consider expanding task coverage and integrating tree-structured generation for richer scenario variations.
%\section*{Acknowledgments}

% This document has been adapted
% by Steven Bethard, Ryan Cotterell and Rui Yan
% from the instructions for earlier ACL and NAACL proceedings, including those for
% ACL 2019 by Douwe Kiela and Ivan Vuli\'{c},
% NAACL 2019 by Stephanie Lukin and Alla Roskovskaya,
% ACL 2018 by Shay Cohen, Kevin Gimpel, and Wei Lu,
% NAACL 2018 by Margaret Mitchell and Stephanie Lukin,
% Bib\TeX{} suggestions for (NA)ACL 2017/2018 from Jason Eisner,
% ACL 2017 by Dan Gildea and Min-Yen Kan,
% NAACL 2017 by Margaret Mitchell,
% ACL 2012 by Maggie Li and Michael White,
% ACL 2010 by Jing-Shin Chang and Philipp Koehn,
% ACL 2008 by Johanna D. Moore, Simone Teufel, James Allan, and Sadaoki Furui,
% ACL 2005 by Hwee Tou Ng and Kemal Oflazer,
% ACL 2002 by Eugene Charniak and Dekang Lin,
% and earlier ACL and EACL formats written by several people, including
% John Chen, Henry S. Thompson and Donald Walker.
% Additional elements were taken from the formatting instructions of the \emph{International Joint Conference on Artificial Intelligence} and the \emph{Conference on Computer Vision and Pattern Recognition}.

\bibliography{custom}

\begin{thebibliography}{71}
\providecommand{\natexlab}[1]{#1}

\bibitem[{Achiam et~al.(2023)Achiam, Adler, Agarwal, Ahmad, Akkaya, Aleman, Almeida, Altenschmidt, Altman, Anadkat et~al.}]{achiam2023gpt}
Josh Achiam, Steven Adler, Sandhini Agarwal, Lama Ahmad, Ilge Akkaya, Florencia~Leoni Aleman, Diogo Almeida, Janko Altenschmidt, Sam Altman, Shyamal Anadkat, et~al. 2023.
\newblock Gpt-4 technical report.
\newblock \emph{arXiv preprint arXiv:2303.08774}.

\bibitem[{Azerbayev et~al.()Azerbayev, Schoelkopf, Paster, Dos~Santos, McAleer, Jiang, Deng, Biderman, and Welleck}]{azerbayevllemma}
Zhangir Azerbayev, Hailey Schoelkopf, Keiran Paster, Marco Dos~Santos, Stephen~Marcus McAleer, Albert~Q Jiang, Jia Deng, Stella Biderman, and Sean Welleck.
\newblock Llemma: An open language model for mathematics.
\newblock In \emph{The Twelfth International Conference on Learning Representations}.

\bibitem[{Chang et~al.(2024)Chang, Wang, Wang, Wu, Yang, Zhu, Chen, Yi, Wang, Wang et~al.}]{chang2024survey}
Yupeng Chang, Xu~Wang, Jindong Wang, Yuan Wu, Linyi Yang, Kaijie Zhu, Hao Chen, Xiaoyuan Yi, Cunxiang Wang, Yidong Wang, et~al. 2024.
\newblock A survey on evaluation of large language models.
\newblock \emph{ACM Transactions on Intelligent Systems and Technology}, 15(3):1--45.

\bibitem[{Cobbe et~al.(2021)Cobbe, Kosaraju, Bavarian, Chen, Jun, Kaiser, Plappert, Tworek, Hilton, Nakano et~al.}]{cobbe2021gsm8k}
Karl Cobbe, Vineet Kosaraju, Mohammad Bavarian, Mark Chen, Heewoo Jun, Lukasz Kaiser, Matthias Plappert, Jerry Tworek, Jacob Hilton, Reiichiro Nakano, et~al. 2021.
\newblock Training verifiers to solve math word problems.
\newblock \emph{arXiv preprint arXiv:2110.14168}.

\bibitem[{Delaney et~al.(2021)Delaney, Greene, and Keane}]{delaney2021uncertainty}
Eoin Delaney, Derek Greene, and Mark~T Keane. 2021.
\newblock Uncertainty estimation and out-of-distribution detection for counterfactual explanations: Pitfalls and solutions.
\newblock \emph{arXiv preprint arXiv:2107.09734}.

\bibitem[{Dubey et~al.(2024)Dubey, Jauhri, Pandey, Kadian, Al-Dahle, Letman, Mathur, Schelten, Yang, Fan et~al.}]{dubey2024llama}
Abhimanyu Dubey, Abhinav Jauhri, Abhinav Pandey, Abhishek Kadian, Ahmad Al-Dahle, Aiesha Letman, Akhil Mathur, Alan Schelten, Amy Yang, Angela Fan, et~al. 2024.
\newblock The llama 3 herd of models.
\newblock \emph{arXiv preprint arXiv:2407.21783}.

\bibitem[{Gao et~al.(2023)Gao, Ilharco, Lundberg, and Ribeiro}]{gao2023adaptive}
Irena Gao, Gabriel Ilharco, Scott Lundberg, and Marco~Tulio Ribeiro. 2023.
\newblock Adaptive testing of computer vision models.
\newblock In \emph{Proceedings of the IEEE/CVF International Conference on Computer Vision}, pages 4003--4014.

\bibitem[{Glazer et~al.(2024)Glazer, Erdil, Besiroglu, Chicharro, Chen, Gunning, Falkman~Olsson, Denain, Ho, de~Oliveira~Santos et~al.}]{glazer2024frontiermath}
Elliot Glazer, Ege Erdil, Tamay Besiroglu, Diego Chicharro, Evan Chen, Alex Gunning, Caroline Falkman~Olsson, Jean-Stanislas Denain, Anson Ho, Emily de~Oliveira~Santos, et~al. 2024.
\newblock Frontiermath: A benchmark for evaluating advanced mathematical reasoning in ai.
\newblock \emph{arXiv e-prints}, pages arXiv--2411.

\bibitem[{Gui et~al.(2024)Gui, Liu, Cheng, Gu, Liu, Wang, Dong, Tang, and Huang}]{gui2024logicgame}
Jiayi Gui, Yiming Liu, Jiale Cheng, Xiaotao Gu, Xiao Liu, Hongning Wang, Yuxiao Dong, Jie Tang, and Minlie Huang. 2024.
\newblock Logicgame: Benchmarking rule-based reasoning abilities of large language models.
\newblock \emph{arXiv preprint arXiv:2408.15778}.

\bibitem[{Guo et~al.(2025)Guo, Yang, Zhang, Song, Zhang, Xu, Zhu, Ma, Wang, Bi et~al.}]{guo2025deepseek}
Daya Guo, Dejian Yang, Haowei Zhang, Junxiao Song, Ruoyu Zhang, Runxin Xu, Qihao Zhu, Shirong Ma, Peiyi Wang, Xiao Bi, et~al. 2025.
\newblock Deepseek-r1: Incentivizing reasoning capability in llms via reinforcement learning.
\newblock \emph{arXiv preprint arXiv:2501.12948}.

\bibitem[{Hao et~al.(2023)Hao, Gu, Ma, Hong, Wang, Wang, and Hu}]{hao2023reasoning}
Shibo Hao, Yi~Gu, Haodi Ma, Joshua~Jiahua Hong, Zhen Wang, Daisy~Zhe Wang, and Zhiting Hu. 2023.
\newblock Reasoning with language model is planning with world model.
\newblock \emph{arXiv preprint arXiv:2305.14992}.

\bibitem[{Hendrycks et~al.(2020)Hendrycks, Burns, Basart, Zou, Mazeika, Song, and Steinhardt}]{hendrycks2020measuring}
Dan Hendrycks, Collin Burns, Steven Basart, Andy Zou, Mantas Mazeika, Dawn Song, and Jacob Steinhardt. 2020.
\newblock Measuring massive multitask language understanding.
\newblock \emph{arXiv preprint arXiv:2009.03300}.

\bibitem[{Hendrycks et~al.(2021)Hendrycks, Burns, Kadavath, Arora, Basart, Tang, Song, and Steinhardt}]{hendrycks2021measuring}
Dan Hendrycks, Collin Burns, Saurav Kadavath, Akul Arora, Steven Basart, Eric Tang, Dawn Song, and Jacob Steinhardt. 2021.
\newblock Measuring mathematical problem solving with the math dataset.
\newblock \emph{arXiv preprint arXiv:2103.03874}.

\bibitem[{Hosseini et~al.(2024)Hosseini, Sordoni, Toyama, Courville, and Agarwal}]{hosseini2024not}
Arian Hosseini, Alessandro Sordoni, Daniel Toyama, Aaron Courville, and Rishabh Agarwal. 2024.
\newblock Not all llm reasoners are created equal.
\newblock \emph{arXiv preprint arXiv:2410.01748}.

\bibitem[{Huang et~al.(2024{\natexlab{a}})Huang, Ma, Li, Huang, Zou, Zhang, and Zheng}]{huang2024lateval}
Shulin Huang, Shirong Ma, Yinghui Li, Mengzuo Huang, Wuhe Zou, Weidong Zhang, and Haitao Zheng. 2024{\natexlab{a}}.
\newblock Lateval: An interactive llms evaluation benchmark with incomplete information from lateral thinking puzzles.
\newblock In \emph{Proceedings of the 2024 Joint International Conference on Computational Linguistics, Language Resources and Evaluation (LREC-COLING 2024)}, pages 10186--10197.

\bibitem[{Huang et~al.(2024{\natexlab{b}})Huang, Bai, Zhu, Zhang, Zhang, Su, Liu, Lv, Zhang, Fu et~al.}]{huang2024c}
Yuzhen Huang, Yuzhuo Bai, Zhihao Zhu, Junlei Zhang, Jinghan Zhang, Tangjun Su, Junteng Liu, Chuancheng Lv, Yikai Zhang, Yao Fu, et~al. 2024{\natexlab{b}}.
\newblock C-eval: A multi-level multi-discipline chinese evaluation suite for foundation models.
\newblock \emph{Advances in Neural Information Processing Systems}, 36.

\bibitem[{Huang et~al.(2024{\natexlab{c}})Huang, Zou, Li, Liu, Zheng, Chern, Xia, Qin, Yuan, and Liu}]{huang2024o1}
Zhen Huang, Haoyang Zou, Xuefeng Li, Yixiu Liu, Yuxiang Zheng, Ethan Chern, Shijie Xia, Yiwei Qin, Weizhe Yuan, and Pengfei Liu. 2024{\natexlab{c}}.
\newblock O1 replication journey--part 2: Surpassing o1-preview through simple distillation, big progress or bitter lesson?
\newblock \emph{arXiv preprint arXiv:2411.16489}.

\bibitem[{Jiang et~al.(2023)Jiang, Sablayrolles, Mensch, Bamford, Chaplot, Casas, Bressand, Lengyel, Lample, Saulnier et~al.}]{jiang2023mistral}
Albert~Q Jiang, Alexandre Sablayrolles, Arthur Mensch, Chris Bamford, Devendra~Singh Chaplot, Diego de~las Casas, Florian Bressand, Gianna Lengyel, Guillaume Lample, Lucile Saulnier, et~al. 2023.
\newblock Mistral 7b.
\newblock \emph{arXiv preprint arXiv:2310.06825}.

\bibitem[{Jiang et~al.(2024)Jiang, Sablayrolles, Roux, Mensch, Savary, Bamford, Chaplot, Casas, Hanna, Bressand et~al.}]{jiang2024mixtral}
Albert~Q Jiang, Alexandre Sablayrolles, Antoine Roux, Arthur Mensch, Blanche Savary, Chris Bamford, Devendra~Singh Chaplot, Diego de~las Casas, Emma~Bou Hanna, Florian Bressand, et~al. 2024.
\newblock Mixtral of experts.
\newblock \emph{arXiv preprint arXiv:2401.04088}.

\bibitem[{Karanikolas et~al.(2023)Karanikolas, Manga, Samaridi, Tousidou, and Vassilakopoulos}]{karanikolas2023large}
Nikitas Karanikolas, Eirini Manga, Nikoletta Samaridi, Eleni Tousidou, and Michael Vassilakopoulos. 2023.
\newblock Large language models versus natural language understanding and generation.
\newblock In \emph{Proceedings of the 27th Pan-Hellenic Conference on Progress in Computing and Informatics}, pages 278--290.

\bibitem[{Kenny and Keane(2021)}]{kenny2021generating}
Eoin~M Kenny and Mark~T Keane. 2021.
\newblock On generating plausible counterfactual and semi-factual explanations for deep learning.

\bibitem[{Lewkowycz et~al.(2022)Lewkowycz, Andreassen, Dohan, Dyer, Michalewski, Ramasesh, Slone, Anil, Schlag, Gutman-Solo et~al.}]{lewkowycz2022solving}
Aitor Lewkowycz, Anders Andreassen, David Dohan, Ethan Dyer, Henryk Michalewski, Vinay Ramasesh, Ambrose Slone, Cem Anil, Imanol Schlag, Theo Gutman-Solo, et~al. 2022.
\newblock Solving quantitative reasoning problems with language models.
\newblock \emph{Advances in Neural Information Processing Systems}, 35:3843--3857.

\bibitem[{Li et~al.(2019)Li, Ji, Du, Li, and Wang}]{li2019textbugger}
J~Li, S~Ji, T~Du, B~Li, and T~Wang. 2019.
\newblock Textbugger: Generating adversarial text against real-world applications.
\newblock In \emph{26th Annual Network and Distributed System Security Symposium}.

\bibitem[{Li et~al.(2024{\natexlab{a}})Li, Luo, and Pan}]{li2024openai}
Leo Li, Ye~Luo, and Tingyou Pan. 2024{\natexlab{a}}.
\newblock Openai-o1 ab testing: Does the o1 model really do good reasoning in math problem solving?
\newblock \emph{arXiv preprint arXiv:2411.06198}.

\bibitem[{Li et~al.(2024{\natexlab{b}})Li, Cui, Zhao, Kong, and Bi}]{li2024gsm}
Qintong Li, Leyang Cui, Xueliang Zhao, Lingpeng Kong, and Wei Bi. 2024{\natexlab{b}}.
\newblock Gsm-plus: A comprehensive benchmark for evaluating the robustness of llms as mathematical problem solvers.
\newblock \emph{arXiv preprint arXiv:2402.19255}.

\bibitem[{Li et~al.(2023)Li, Zhang, Dubois, Taori, Gulrajani, Guestrin, Liang, and Hashimoto}]{li2023alpacaeval}
Xuechen Li, Tianyi Zhang, Yann Dubois, Rohan Taori, Ishaan Gulrajani, Carlos Guestrin, Percy Liang, and Tatsunori~B Hashimoto. 2023.
\newblock Alpacaeval: An automatic evaluator of instruction-following models.

\bibitem[{Li et~al.(2022)Li, Lin, Zhang, Fu, Chen, Lou, and Chen}]{li2022making}
Yifei Li, Zeqi Lin, Shizhuo Zhang, Qiang Fu, Bei Chen, Jian-Guang Lou, and Weizhu Chen. 2022.
\newblock Making large language models better reasoners with step-aware verifier.
\newblock \emph{arXiv preprint arXiv:2206.02336}.

\bibitem[{Li et~al.(2024{\natexlab{c}})Li, Zhou, Luo, Ma, Li, Zheng, Hu, and Yu}]{li2024llms}
Yinghui Li, Qingyu Zhou, Yuanzhen Luo, Shirong Ma, Yangning Li, Hai-Tao Zheng, Xuming Hu, and Philip~S Yu. 2024{\natexlab{c}}.
\newblock When llms meet cunning questions: A fallacy understanding benchmark for large language models.
\newblock \emph{arXiv preprint arXiv:2402.11100}.

\bibitem[{Liang et~al.(2022)Liang, Bommasani, Lee, Tsipras, Soylu, Yasunaga, Zhang, Narayanan, Wu, Kumar et~al.}]{liang2022holistic}
Percy Liang, Rishi Bommasani, Tony Lee, Dimitris Tsipras, Dilara Soylu, Michihiro Yasunaga, Yian Zhang, Deepak Narayanan, Yuhuai Wu, Ananya Kumar, et~al. 2022.
\newblock Holistic evaluation of language models.
\newblock \emph{arXiv preprint arXiv:2211.09110}.

\bibitem[{Lightman et~al.(2023)Lightman, Kosaraju, Burda, Edwards, Baker, Lee, Leike, Schulman, Sutskever, and Cobbe}]{lightman2023let}
Hunter Lightman, Vineet Kosaraju, Yura Burda, Harri Edwards, Bowen Baker, Teddy Lee, Jan Leike, John Schulman, Ilya Sutskever, and Karl Cobbe. 2023.
\newblock Let's verify step by step.
\newblock \emph{arXiv preprint arXiv:2305.20050}.

\bibitem[{Lin et~al.(2024)Lin, Bras, and Choi}]{zebralogic2024}
Bill~Yuchen Lin, Ronan~Le Bras, and Yejin Choi. 2024.
\newblock \href {https://huggingface.co/spaces/allenai/ZebraLogic} {Zebralogic: Benchmarking the logical reasoning ability of language models}.

\bibitem[{Liu et~al.(2024)Liu, Feng, Xue, Wang, Wu, Lu, Zhao, Deng, Zhang, Ruan et~al.}]{liu2024deepseek}
Aixin Liu, Bei Feng, Bing Xue, Bingxuan Wang, Bochao Wu, Chengda Lu, Chenggang Zhao, Chengqi Deng, Chenyu Zhang, Chong Ruan, et~al. 2024.
\newblock Deepseek-v3 technical report.
\newblock \emph{arXiv preprint arXiv:2412.19437}.

\bibitem[{Liu et~al.(2025)Liu, Gao, Zhao, Zhang, Li, Qi, Ouyang, and Zhou}]{Liu2025Can1L}
Runze Liu, Junqi Gao, Jian Zhao, Kaiyan Zhang, Xiu Li, Biqing Qi, Wanli Ouyang, and Bowen Zhou. 2025.
\newblock \href {https://api.semanticscholar.org/CorpusID:276249339} {Can 1b llm surpass 405b llm? rethinking compute-optimal test-time scaling}.

\bibitem[{Lu et~al.(2023)Lu, Bansal, Xia, Liu, Li, Hajishirzi, Cheng, Chang, Galley, and Gao}]{lu23mathvista}
Pan Lu, Hritik Bansal, Tony Xia, Jiacheng Liu, Chunyuan Li, Hannaneh Hajishirzi, Hao Cheng, Kai-Wei Chang, Michel Galley, and Jianfeng Gao. 2023.
\newblock Mathvista: Evaluating mathematical reasoning of foundation models in visual contexts.
\newblock In \emph{The 3rd Workshop on Mathematical Reasoning and AI at NeurIPS'23}.

\bibitem[{Luo et~al.(2025)Luo, Tan, Wong, Shi, Tang, Roongta, Cai, Luo, Zhang, Li, Popa, and Stoica}]{deepscaler2025}
Michael Luo, Sijun Tan, Justin Wong, Xiaoxiang Shi, William Tang, Manan Roongta, Colin Cai, Jeffrey Luo, Tianjun Zhang, Erran Li, Raluca~Ada Popa, and Ion Stoica. 2025.
\newblock Deepscaler: Surpassing o1-preview with a 1.5b model by scaling rl.
\newblock Notion Blog.

\bibitem[{Muennighoff et~al.(2025)Muennighoff, Yang, Shi, Li, Fei-Fei, Hajishirzi, Zettlemoyer, Liang, Cand{\`e}s, and Hashimoto}]{muennighoff2025s1}
Niklas Muennighoff, Zitong Yang, Weijia Shi, Xiang~Lisa Li, Li~Fei-Fei, Hannaneh Hajishirzi, Luke Zettlemoyer, Percy Liang, Emmanuel Cand{\`e}s, and Tatsunori Hashimoto. 2025.
\newblock s1: Simple test-time scaling.
\newblock \emph{arXiv preprint arXiv:2501.19393}.

\bibitem[{Naik et~al.(2018)Naik, Ravichander, Sadeh, Rose, and Neubig}]{naik2018stress}
Aakanksha Naik, Abhilasha Ravichander, Norman Sadeh, Carolyn Rose, and Graham Neubig. 2018.
\newblock Stress test evaluation for natural language inference.
\newblock In \emph{Proceedings of the 27th International Conference on Computational Linguistics}, pages 2340--2353.

\bibitem[{Naveed et~al.(2023)Naveed, Khan, Qiu, Saqib, Anwar, Usman, Akhtar, Barnes, and Mian}]{naveed2023comprehensive}
Humza Naveed, Asad~Ullah Khan, Shi Qiu, Muhammad Saqib, Saeed Anwar, Muhammad Usman, Naveed Akhtar, Nick Barnes, and Ajmal Mian. 2023.
\newblock A comprehensive overview of large language models.
\newblock \emph{arXiv preprint arXiv:2307.06435}.

\bibitem[{o1~Team(2024)}]{skyworkopeno12024}
Skywork o1~Team. 2024.
\newblock \href {https://huggingface.co/Skywork} {Skywork-o1 open series}.
\newblock \url{https://huggingface.co/Skywork}.

\bibitem[{Opedal et~al.(2024)Opedal, Shirakami, Sch{\"o}lkopf, Saparov, and Sachan}]{opedal2024mathgap}
Andreas Opedal, Haruki Shirakami, Bernhard Sch{\"o}lkopf, Abulhair Saparov, and Mrinmaya Sachan. 2024.
\newblock Mathgap: Out-of-distribution evaluation on problems with arbitrarily complex proofs.
\newblock \emph{arXiv preprint arXiv:2410.13502}.

\bibitem[{OpenAI(2024)}]{o1}
OpenAI. 2024.
\newblock Learning to reason with llms.
\newblock \url{https://openai.com/index/learning-to-reason-with-llms/ }.

\bibitem[{OpenAI(2025)}]{o3}
OpenAI. 2025.
\newblock Openai o3-mini.
\newblock \url{https://openai.com/index/openai-o3-mini/ }.

\bibitem[{Qin et~al.(2024)Qin, Li, Zou, Liu, Xia, Huang, Ye, Yuan, Liu, Li, and Liu}]{o1journey}
Yiwei Qin, Xuefeng Li, Haoyang Zou, Yixiu Liu, Shijie Xia, Zhen Huang, Yixin Ye, Weizhe Yuan, Zhengzhong Liu, Yuanzhi Li, and Pengfei Liu. 2024.
\newblock O1 replication journey: A strategic progress report – part 1.
\newblock \url{https://github.com/GAIR-NLP/O1-Journey}.

\bibitem[{Rein et~al.(2023)Rein, Hou, Stickland, Petty, Pang, Dirani, Michael, and Bowman}]{rein2023gpqa}
David Rein, Betty~Li Hou, Asa~Cooper Stickland, Jackson Petty, Richard~Yuanzhe Pang, Julien Dirani, Julian Michael, and Samuel~R Bowman. 2023.
\newblock Gpqa: A graduate-level google-proof q\&a benchmark.
\newblock \emph{arXiv preprint arXiv:2311.12022}.

\bibitem[{Ribeiro and Lundberg(2022)}]{ribeiro2022adaptive}
Marco~Tulio Ribeiro and Scott Lundberg. 2022.
\newblock Adaptive testing and debugging of nlp models.
\newblock In \emph{Proceedings of the 60th Annual Meeting of the Association for Computational Linguistics (Volume 1: Long Papers)}, pages 3253--3267.

\bibitem[{Ribeiro et~al.(2020)Ribeiro, Wu, Guestrin, and Singh}]{ribeiro2020beyond}
Marco~Tulio Ribeiro, Tongshuang Wu, Carlos Guestrin, and Sameer Singh. 2020.
\newblock Beyond accuracy: Behavioral testing of nlp models with checklist.
\newblock In \emph{Proceedings of the 58th Annual Meeting of the Association for Computational Linguistics}, pages 4902--4912.

\bibitem[{Rivi{\`e}re et~al.(2024)Rivi{\`e}re, Pathak, Sessa, Hardin, Bhupatiraju, Hussenot, Mesnard, Shahriari, Ram{\'e}, Ferret et~al.}]{riviere2024gemma}
Morgane Rivi{\`e}re, Shreya Pathak, Pier~Giuseppe Sessa, Cassidy Hardin, Surya Bhupatiraju, L{\'e}onard Hussenot, Thomas Mesnard, Bobak Shahriari, Alexandre Ram{\'e}, Johan Ferret, et~al. 2024.
\newblock Gemma 2: Improving open language models at a practical size.
\newblock \emph{CoRR}.

\bibitem[{Uesato et~al.(2022)Uesato, Kushman, Kumar, Song, Siegel, Wang, Creswell, Irving, and Higgins}]{uesato2022solving}
Jonathan Uesato, Nate Kushman, Ramana Kumar, Francis Song, Noah Siegel, Lisa Wang, Antonia Creswell, Geoffrey Irving, and Irina Higgins. 2022.
\newblock Solving math word problems with process-and outcome-based feedback.
\newblock \emph{arXiv preprint arXiv:2211.14275}.

\bibitem[{Wang et~al.(2023{\natexlab{a}})Wang, Hu, Hou, Chen, Zheng, Wang, Yang, Huang, Ye, Geng et~al.}]{wang2023robustness}
Jindong Wang, Xixu Hu, Wenxin Hou, Hao Chen, Runkai Zheng, Yidong Wang, Linyi Yang, Haojun Huang, Wei Ye, Xiubo Geng, et~al. 2023{\natexlab{a}}.
\newblock On the robustness of chatgpt: An adversarial and out-of-distribution perspective.
\newblock \emph{arXiv preprint arXiv:2302.12095}.

\bibitem[{Wang et~al.(2024{\natexlab{a}})Wang, Fang, Wan, Wen, Zhu, Liu, Gong, Song, Chen, Ni et~al.}]{wang2024openr}
Jun Wang, Meng Fang, Ziyu Wan, Muning Wen, Jiachen Zhu, Anjie Liu, Ziqin Gong, Yan Song, Lei Chen, Lionel~M Ni, et~al. 2024{\natexlab{a}}.
\newblock Openr: An open source framework for advanced reasoning with large language models.
\newblock \emph{arXiv preprint arXiv:2410.09671}.

\bibitem[{Wang et~al.(2024{\natexlab{b}})Wang, Li, Bhatt, Xi, Liu, Topcu, and Wang}]{wang2024planning}
Kevin Wang, Junbo Li, Neel~P Bhatt, Yihan Xi, Qiang Liu, Ufuk Topcu, and Zhangyang Wang. 2024{\natexlab{b}}.
\newblock On the planning abilities of openai's o1 models: Feasibility, optimality, and generalizability.
\newblock \emph{arXiv preprint arXiv:2409.19924}.

\bibitem[{Wang et~al.(2024{\natexlab{c}})Wang, Li, Shao, Xu, Dai, Li, Chen, Wu, and Sui}]{wang2024math}
Peiyi Wang, Lei Li, Zhihong Shao, Runxin Xu, Damai Dai, Yifei Li, Deli Chen, Yu~Wu, and Zhifang Sui. 2024{\natexlab{c}}.
\newblock Math-shepherd: Verify and reinforce llms step-by-step without human annotations.
\newblock In \emph{Proceedings of the 62nd Annual Meeting of the Association for Computational Linguistics (Volume 1: Long Papers)}, pages 9426--9439.

\bibitem[{Wang et~al.(2022)Wang, Wei, Schuurmans, Le, Chi, and Zhou}]{wang2022self}
Xuezhi Wang, Jason Wei, Dale Schuurmans, Quoc Le, Ed~Chi, and Denny Zhou. 2022.
\newblock Self-consistency improves chain of thought reasoning in language models.
\newblock \emph{arXiv preprint arXiv:2203.11171}.

\bibitem[{Wang et~al.(2023{\natexlab{b}})Wang, Yu, Zeng, Yang, Wang, Chen, Jiang, Xie, Wang, Xie et~al.}]{wang2023pandalm}
Yidong Wang, Zhuohao Yu, Zhengran Zeng, Linyi Yang, Cunxiang Wang, Hao Chen, Chaoya Jiang, Rui Xie, Jindong Wang, Xing Xie, et~al. 2023{\natexlab{b}}.
\newblock Pandalm: An automatic evaluation benchmark for llm instruction tuning optimization.
\newblock \emph{arXiv preprint arXiv:2306.05087}.

\bibitem[{Wu et~al.(2024{\natexlab{a}})Wu, Yang, Okumura, and Zhang}]{wu2024mrke}
Jian Wu, Linyi Yang, Manabu Okumura, and Yue Zhang. 2024{\natexlab{a}}.
\newblock Mrke: The multi-hop reasoning evaluation of llms by knowledge edition.
\newblock \emph{arXiv preprint arXiv:2402.11924}.

\bibitem[{Wu et~al.(2024{\natexlab{b}})Wu, Bansal, Zhang, Wu, Li, Zhu, Jiang, Zhang, Zhang, Liu et~al.}]{wu2024autogen}
Qingyun Wu, Gagan Bansal, Jieyu Zhang, Yiran Wu, Beibin Li, Erkang Zhu, Li~Jiang, Xiaoyun Zhang, Shaokun Zhang, Jiale Liu, et~al. 2024{\natexlab{b}}.
\newblock Autogen: Enabling next-gen llm applications via multi-agent conversation.
\newblock In \emph{ICLR 2024 Workshop on Large Language Model (LLM) Agents}.

\bibitem[{Xu et~al.(2024)Xu, Chen, Peng, Zhang, Xu, Zhao, Wu, Zheng, Wang, and Chen}]{xu2024large}
Derong Xu, Wei Chen, Wenjun Peng, Chao Zhang, Tong Xu, Xiangyu Zhao, Xian Wu, Yefeng Zheng, Yang Wang, and Enhong Chen. 2024.
\newblock Large language models for generative information extraction: A survey.
\newblock \emph{Frontiers of Computer Science}, 18(6):186357.

\bibitem[{Yang et~al.(2024)Yang, Zhang, Hui, Gao, Yu, Li, Liu, Tu, Zhou, Lin et~al.}]{yang2024qwen2}
An~Yang, Beichen Zhang, Binyuan Hui, Bofei Gao, Bowen Yu, Chengpeng Li, Dayiheng Liu, Jianhong Tu, Jingren Zhou, Junyang Lin, et~al. 2024.
\newblock Qwen2. 5-math technical report: Toward mathematical expert model via self-improvement.
\newblock \emph{arXiv preprint arXiv:2409.12122}.

\bibitem[{Yang et~al.(2022)Yang, Yuan, Cui, Gao, and Zhang}]{yang2022factmix}
Linyi Yang, Lifan Yuan, Leyang Cui, Wenyang Gao, and Yue Zhang. 2022.
\newblock Factmix: Using a few labeled in-domain examples to generalize to cross-domain named entity recognition.
\newblock In \emph{Proceedings of the 29th International Conference on Computational Linguistics}, pages 5360--5371.

\bibitem[{Yang et~al.(2023)Yang, Zhang, Qin, Li, Wang, Liu, Wang, Xie, and Zhang}]{yang2023glue}
Linyi Yang, Shuibai Zhang, Libo Qin, Yafu Li, Yidong Wang, Hanmeng Liu, Jindong Wang, Xing Xie, and Yue Zhang. 2023.
\newblock Glue-x: Evaluating natural language understanding models from an out-of-distribution generalization perspective.
\newblock In \emph{Findings of the Association for Computational Linguistics: ACL 2023}, pages 12731--12750.

\bibitem[{Yu et~al.(2024)Yu, Gao, and Wang}]{yu2024ovm}
Fei Yu, Anningzhe Gao, and Benyou Wang. 2024.
\newblock Ovm, outcome-supervised value models for planning in mathematical reasoning.
\newblock In \emph{Findings of the Association for Computational Linguistics: NAACL 2024}, pages 858--875.

\bibitem[{Yuan et~al.(2023)Yuan, Chen, Cui, Gao, Zou, Cheng, Ji, Liu, and Sun}]{yuan2023revisiting}
Lifan Yuan, Yangyi Chen, Ganqu Cui, Hongcheng Gao, Fangyuan Zou, Xingyi Cheng, Heng Ji, Zhiyuan Liu, and Maosong Sun. 2023.
\newblock Revisiting out-of-distribution robustness in nlp: Benchmarks, analysis, and llms evaluations.
\newblock \emph{Advances in Neural Information Processing Systems}, 36:58478--58507.

\bibitem[{Zhang et~al.(2024)Zhang, Hosseini, Bansal, Kazemi, Kumar, and Agarwal}]{zhang2024generative}
Lunjun Zhang, Arian Hosseini, Hritik Bansal, Mehran Kazemi, Aviral Kumar, and Rishabh Agarwal. 2024.
\newblock Generative verifiers: Reward modeling as next-token prediction.
\newblock \emph{arXiv preprint arXiv:2408.15240}.

\bibitem[{Zhang et~al.(2025)Zhang, Zheng, Wu, Zhang, Lin, Yu, Liu, Zhou, and Lin}]{zhang2025lessons}
Zhenru Zhang, Chujie Zheng, Yangzhen Wu, Beichen Zhang, Runji Lin, Bowen Yu, Dayiheng Liu, Jingren Zhou, and Junyang Lin. 2025.
\newblock The lessons of developing process reward models in mathematical reasoning.
\newblock \emph{arXiv preprint arXiv:2501.07301}.

\bibitem[{Zheng et~al.(2024)Zheng, Zhang, Zhang, Lin, Lu, Yu, Liu, Zhou, and Lin}]{zheng2024processbench}
Chujie Zheng, Zhenru Zhang, Beichen Zhang, Runji Lin, Keming Lu, Bowen Yu, Dayiheng Liu, Jingren Zhou, and Junyang Lin. 2024.
\newblock \href {https://arxiv.org/abs/2412.06559} {Processbench: Identifying process errors in mathematical reasoning}.
\newblock \emph{Preprint}, arXiv:2412.06559.

\bibitem[{Zheng et~al.(2021)Zheng, Han, and Polu}]{zhengminif2f}
Kunhao Zheng, Jesse~Michael Han, and Stanislas Polu. 2021.
\newblock minif2f: a cross-system benchmark for formal olympiad-level mathematics.
\newblock In \emph{International Conference on Learning Representations}.

\bibitem[{Zheng et~al.(2023)Zheng, Chiang, Sheng, Zhuang, Wu, Zhuang, Lin, Li, Li, Xing, Zhang, Gonzalez, and Stoica}]{zheng2023judging}
Lianmin Zheng, Wei-Lin Chiang, Ying Sheng, Siyuan Zhuang, Zhanghao Wu, Yonghao Zhuang, Zi~Lin, Zhuohan Li, Dacheng Li, Eric.~P Xing, Hao Zhang, Joseph~E. Gonzalez, and Ion Stoica. 2023.
\newblock \href {https://arxiv.org/abs/2306.05685} {Judging llm-as-a-judge with mt-bench and chatbot arena}.
\newblock \emph{Preprint}, arXiv:2306.05685.

\bibitem[{Zhong et~al.(2024)Zhong, Liu, Pan, Zhang, Zhou, Liang, Wu, Lyu, Shu, Yu et~al.}]{zhong2024evaluation}
Tianyang Zhong, Zhengliang Liu, Yi~Pan, Yutong Zhang, Yifan Zhou, Shizhe Liang, Zihao Wu, Yanjun Lyu, Peng Shu, Xiaowei Yu, et~al. 2024.
\newblock Evaluation of openai o1: Opportunities and challenges of agi.
\newblock \emph{arXiv preprint arXiv:2409.18486}.

\bibitem[{Zhu et~al.(2023{\natexlab{a}})Zhu, Chen, Wang, Gong, Yang, and Xie}]{zhu2023dyval}
Kaijie Zhu, Jiaao Chen, Jindong Wang, Neil~Zhenqiang Gong, Diyi Yang, and Xing Xie. 2023{\natexlab{a}}.
\newblock Dyval: Dynamic evaluation of large language models for reasoning tasks.
\newblock In \emph{The Twelfth International Conference on Learning Representations}.

\bibitem[{Zhu et~al.(2024)Zhu, Wang, Zhao, Xu, and Xie}]{zhu2024dyval}
Kaijie Zhu, Jindong Wang, Qinlin Zhao, Ruochen Xu, and Xing Xie. 2024.
\newblock Dyval 2: Dynamic evaluation of large language models by meta probing agents.
\newblock \emph{arXiv preprint arXiv:2402.14865}.

\bibitem[{Zhu et~al.(2023{\natexlab{b}})Zhu, Wang, Zhou, Wang, Chen, Wang, Yang, Ye, Zhang, Zhenqiang~Gong et~al.}]{zhu2023promptbench}
Kaijie Zhu, Jindong Wang, Jiaheng Zhou, Zichen Wang, Hao Chen, Yidong Wang, Linyi Yang, Wei Ye, Yue Zhang, Neil Zhenqiang~Gong, et~al. 2023{\natexlab{b}}.
\newblock Promptbench: Towards evaluating the robustness of large language models on adversarial prompts.
\newblock \emph{arXiv e-prints}, pages arXiv--2306.

\end{thebibliography}

\appendix

% \section{Example Appendix}
% \label{sec:appendix}

\newpage
\section{Appendix}

\begin{table*}
 \centering
 \small
\caption{Performance on Scenario-level semi-fact data and Attack-level semi-fact data. Each performance value represents the accuracy in solving problems from the dataset. The accuracy of performance on attack-level semi-factual data is determined by computing the minimum accuracy across three different attacks.}
 % \begin{adjustbox}{width=.7\textwidth}{
\begin{tabular}{lcccccc} 
\toprule
\multirow{2}{*}{}    & \multicolumn{3}{c}{AIME 2024} & \multicolumn{3}{c}{AIME-500}  \\ 
\cmidrule(l){2-7}
                     & Original & OOD (Scenario) & OOD (Attack)  & Original & OOD (Scenario) & OOD (Attack)  \\ 
\midrule
o1-preview           & 0.500    & 0.500    & 0.467   & 0.742    & 0.638    & 0.680   \\
o1-mini              & 0.567    & 0.600    & 0.600   & 0.864    & 0.756    & 0.750   \\
o3-mini              & 0.767    & 0.667    & 0.767   & 0.922    & 0.848    & 0.820   \\
Deepseek-R1          & 0.800    & 0.733    & 0.767   & 0.920    & 0.816    & 0.840   \\
GPT-4o               & 0.133    & 0.100    & 0.100   & 0.278    & 0.204    & 0.190   \\
Deepseek-V3          & 0.367    & 0.333    & 0.333   & 0.528    & 0.438    & 0.420   \\ 
\midrule
Mixtral-8x7B-IT-v0.1 & 0.000    & 0.000    & 0.000   & 0.012    & 0.000    & 0.012   \\
Qwen2.5-72B-IT       & 0.200    & 0.167    & 0.200   & 0.432    & 0.290    & 0.392   \\
Qwen2.5-Math-72B-IT  & 0.267    & 0.233    & 0.200   & 0.536    & 0.360    & 0.500   \\
LLAMA3.1-70B-IT      & 0.200    & 0.167    & 0.167   & 0.424    & 0.244    & 0.392   \\ 
\midrule
s1.1-32B             & 0.533    & 0.500    & 0.478   & 0.752    & 0.654    & 0.494   \\
Gemma-2-27B-IT       & 0.033    & 0.033    & 0.000   & 0.062    & 0.028    & 0.062   \\
Gemma-2-9B-IT        & 0.000    & 0.000    & 0.000   & 0.032    & 0.016    & 0.026   \\
LLAMA3.1-8B-IT       & 0.000    & 0.033    & 0.000   & 0.132    & 0.074    & 0.100   \\
% Qwen2.5-Math-7B      & 0.067    & 0.100    & 0.100   & 0.426    & 0.334    & 0.382   \\
Phi-3-mini-4k-IT     & 0.000    & 0.000    & 0.000   & 0.046    & 0.024    & 0.028   \\
LLAMA3.2-3B-IT       & 0.033    & 0.033    & 0.033   & 0.122    & 0.066    & 0.108   \\
\bottomrule
\end{tabular}
% }
%     \end{adjustbox}
        \label{tab:main table2}
\end{table*}

\subsection{Process Reward Models}
In the past, language model training primarily used Outcome-based Reinforcement Models (ORM)~\citep{wang2024openr}. A foundational example is the ORM-based model by~\citet{cobbe2021gsm8k}, which focuses on training evaluators to assess the correctness of answers, providing crucial feedback. Meanwhile, the Process Reward Model (PRM) aims to provide stepwise rewards, offering fine-grained supervision.
DeepMind~\citep{uesato2022solving} supervises both reasoning steps and final results, while OpenAI~\citep{lightman2023let} introduces PRM800K, a human-annotated dataset, emphasizing step verification. \citet{li2022making} enhance result reliability with evaluator models and majority voting. \citet{yu2024ovm} improve reasoning through reinforcement learning with outcome and process supervision. The Generative Reward Model (GenRM)\citep{zhang2024generative} allows rich interaction between evaluators and generators, reflecting a demand for sophisticated process supervision. Recent work~\citep{zheng2024processbench} offers a benchmark for evaluating error identification in mathematical reasoning, fostering scalable oversight research. As a supplement, we propose a dynamic benchmark for testing reasoning capability in LLMs with the help of PRMs.

\subsection{Process Reward Models on Test-time Computation Budget}
\subsubsection{Settings}
% To comprehensively compare the impact of different PRM (Performance Reward Model) methods on model performance
To comprehensively conduct the test-time scaling evaluation for the various PRMs on AIME 2024 and AIME-500, we employ these PRMs: 

(1) \textbf{Math-Shepherd-PRM}: 
The scoring mechanism employed in Math-Shepherd is essential for understanding the performance of reasoning steps. In this framework, the token ``kn'' is used to indicate the position where the step score is predicted. A ``+'' token represents a good step, one that contributes positively towards reaching the correct answer, while a ``-'' token signals a bad step. Notably, during the training of PRMs, the loss is computed only at positions marked with ``kn''. The PRM denoted as \( (P \times S \to \mathbb{R}^+) \), assigns a score to each reasoning step \( s \). The model is typically trained using the following binary cross-entropy loss function:
\begin{equation}
\mathcal{L}_{PRM} = \sum_{i=1}^{K} y_{s_i} \log r_{s_i} + (1 - y_{s_i}) \log (1 - r_{s_i}),
\end{equation}

where \( y_{s_i} \) represents the ground-truth label of the \( i \)-th reasoning step \( s_i \), \( r_{s_i} \) is the sigmoid output score predicted by the PRM for step \( s_i \), and \( K \) is the total number of reasoning steps in a given solution \( s \). To estimate the quality of each reasoning step, two methods are employed: hard estimation (HE) and soft estimation (SE). In HE, a step is considered good if it contributes to reaching the correct answer \( a^* \), as defined by:

\begin{equation}
y^{HE}_{s_i} =
\begin{cases} 
1, & \text{if } \exists a_j \in A, a_j = a^* \\
0, & \text{otherwise}.
\end{cases}
\end{equation}

In contrast, SE estimates the quality of a step based on the frequency with which it leads to the correct answer across multiple attempts:

\begin{equation}
y^{SE}_{s_i} = \frac{\sum_{j=1}^{N} \mathbb{I}(a_j = a^*)}{N},
\end{equation}

where \( \mathbb{I}(a_j = a^*) \) is an indicator function that returns 1 if the \( j \)-th attempt \( a_j \) equals the correct answer \( a^* \), and \( N \) is the total number of attempts. Once the labels for each reasoning step are obtained, the PRM is trained using the cross-entropy loss function to optimize the model’s ability to correctly classify each step as either good or bad (\cite{wang2024math}).

(2) \textbf{OpenR-PRM}:
The Problem Resolution Model (PRM) computes a score, denoted as \( p_t \), based on the current problem \( q \) and the sequence of solution steps up to time \( t \), represented as \( [x_1, \dots, x_t] \). This approach allows for a precise and detailed analysis of the solution process, helping identify errors as they occur, as noted by \cite{lightman2023let}. The main goal of PRMs is to assess whether the solution process is on track, with a score \( y_t \) calculated to represent the correctness of the solution at step \( t \), ranging from 0 to 1. This score is given by \( y_t = \text{PRM}(q, x_1, x_2, \dots, x_t) \). During training, the model is framed as a next-token prediction task, where labels are assigned as correct or incorrect, represented by positive (+) or negative (-) tokens. After training, OpenR uses the PRM to evaluate the correctness of each solution step during inference, assigning a score \( r_{PRM_t} \) to each step. Two strategies are employed to compute a final score: the PRM-Min strategy, which selects the minimum score among all steps, \( v = \min\{r_{PRM_t}\}_{t=0}^T \), and the PRM-Last strategy, which takes the score of the final step, \( v = r_{PRM_T} \). PRMs function as dense verifiers, providing strong feedback that can significantly enhance the overall solution process (\cite{wang2024openr}).

(3) \textbf{Skywork-PRM}: We evaluate Skywork o1 Open-PRM-Qwen-2.5-7B~\cite{skyworkopeno12024}, which Extends the capabilities of the 1.5B model by scaling up to handle more demanding reasoning tasks, pushing the boundaries of AI reasoning.

(4) \textbf{Qwen-PRM}: We evaluate 7B version of Qwen-PRM~\cite{zhang2025lessons}. This is a process reward model designed to offer feedback on the quality of reasoning and intermediate steps in mathematical problems. It is part of the Qwen2.5-Math series and has shown impressive performance in identifying errors in reasoning processes.

As shown in Figure \ref{fig:analysis}, the model's performance improves on the AIME 2024 OOD and AIME-500 OOD dataset as the test-time computation budget increases and the best-of-N performance of each reward model varies as a function of N. Since majority voting is known to be a strong baseline(\cite{lewkowycz2022solving, wang2022self}), we also include this method as a point of comparison. We compared a series of PRM methods and then observed the performance changes of various PRM methods as N increased. 
% All four series of PRM methods are more effective than majority voting.

% Let's focus more closely on the figure \ref{fig:analysis}. All methods demonstrate an improvement in problem-solving performance as the test-time computation budget $N$ increases, corroborating the Test-time Scaling Law. 
% While Skywork-PRM achieves the best performance across all computation budgets, showing a steepeN values. PRM (o1journey) closely follows Math-Shepherd and consistently outperforms the remaining methods and PRM (OpenR) shows moderate performance, improving steadily with increasing N, but its improvement rate is slower compared to Math-Shepherd and PRM(o1journey).GenRM achieves lower performance, with improvements being less pronounced compared to top-performing methods. Besides, Majority Voting consistently underperforms all other methods, even with increased computation budgets, showing limited scalability.

Skywork-PRM and Qwen-PRM achieve relatively high performance across all computation budgets. OpenR shows moderate performance, improving steadily with increasing N, but its improvement rate is slower compared to Skywork-PRM and Qwen-PRM. Majority Voting mainly underperforms other methods, even with increased computation budgets, showing limited scalability.
The possible underlying reasons are these: (1)Model-specific Capabilities: The superior performance of Skywork-PRM and Qwen-PRM may result from their advanced mechanisms for aggregating or utilizing the increased number of solutions, likely benefiting from better exploration of the solution space or more robust voting schemes. (2)Algorithmic Limitations: The suboptimal performance of Majority Voting suggests its inability to effectively utilize additional solutions. 
% For Majority Voting, this may stem from the lack of sophisticated reasoning mechanisms, while GenRM might suffer from insufficient diversity or quality in the generated solutions. 
(3)Dataset Complexity: The AIME 2024 OOD dataset and AIME-500 OOD dataset likely require nuanced reasoning and adaptability, favoring methods like Skywork-PRM and Qwen-PRM that can better handle out-of-distribution (OOD) generalization tasks. Thus, the results highlight the importance of adopting advanced techniques that can efficiently utilize increased test-time computation budgets. Methods like Skywork-PRM and Qwen-PRM demonstrate superior scaling behavior, suggesting their robustness and adaptability in solving complex problems. 
% In contrast, traditional approaches such as Majority Voting exhibit limited scalability, likely due to its inability to fully exploit the increased computation resources.

% \begin{figure}[t]
%     \centering
%     \includegraphics[width=0.95\columnwidth]{Graph/gpqa_2.pdf}
%     \caption{
%     The performance gap between OOD and ID test on GPQA Diamond. 
%     }
%     \label{fig:gpqa_bar} 
% \end{figure}

\begin{figure}[t]
    \centering
    \includegraphics[width=0.95\columnwidth]{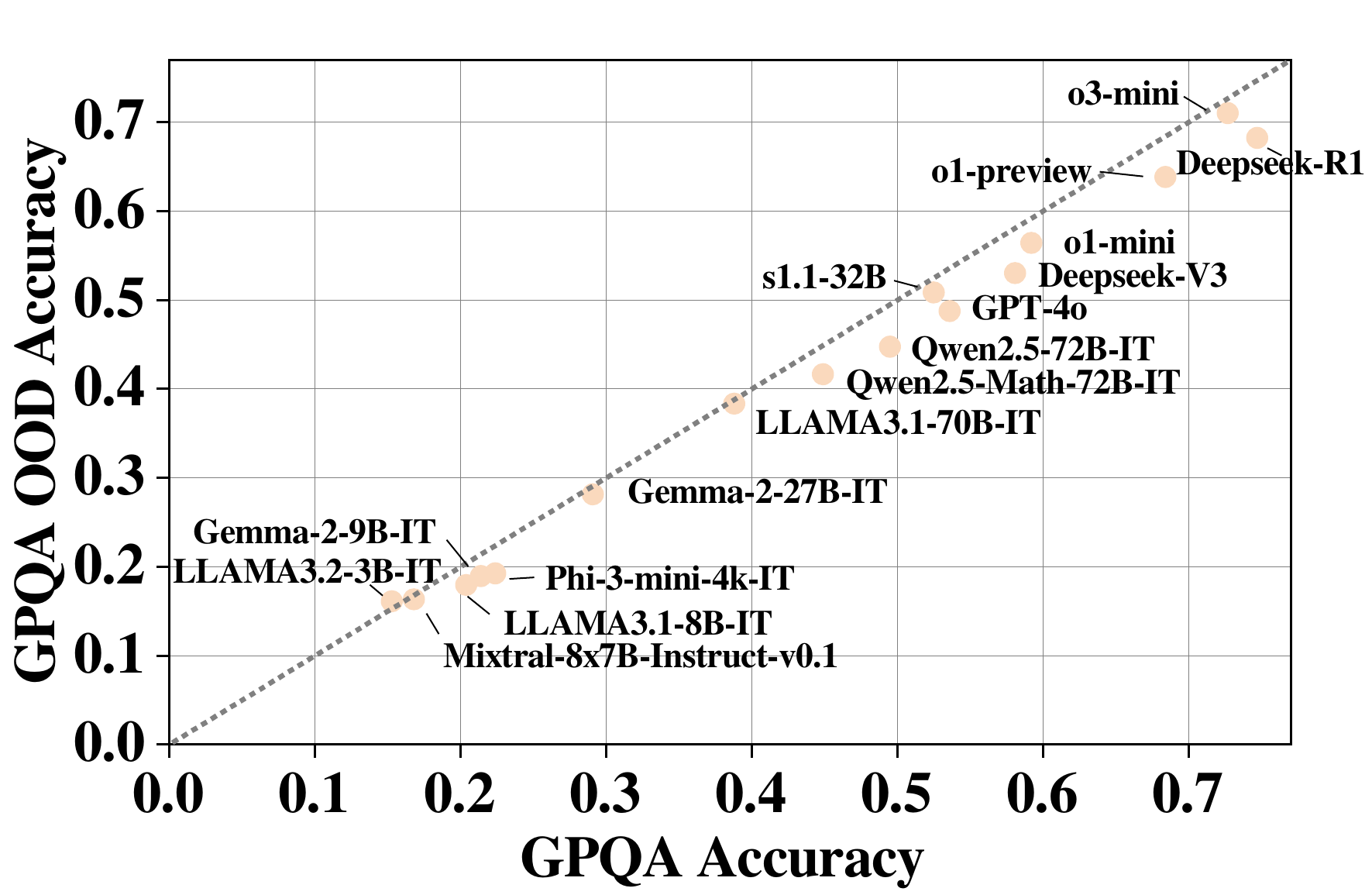}
    \caption{
   Scientific Questions Gap on GPQA Diamond.
    }
    \label{fig:gpqa_line}
\end{figure}

\begin{figure}[t]
    \centering
    \includegraphics[width=0.95\columnwidth]{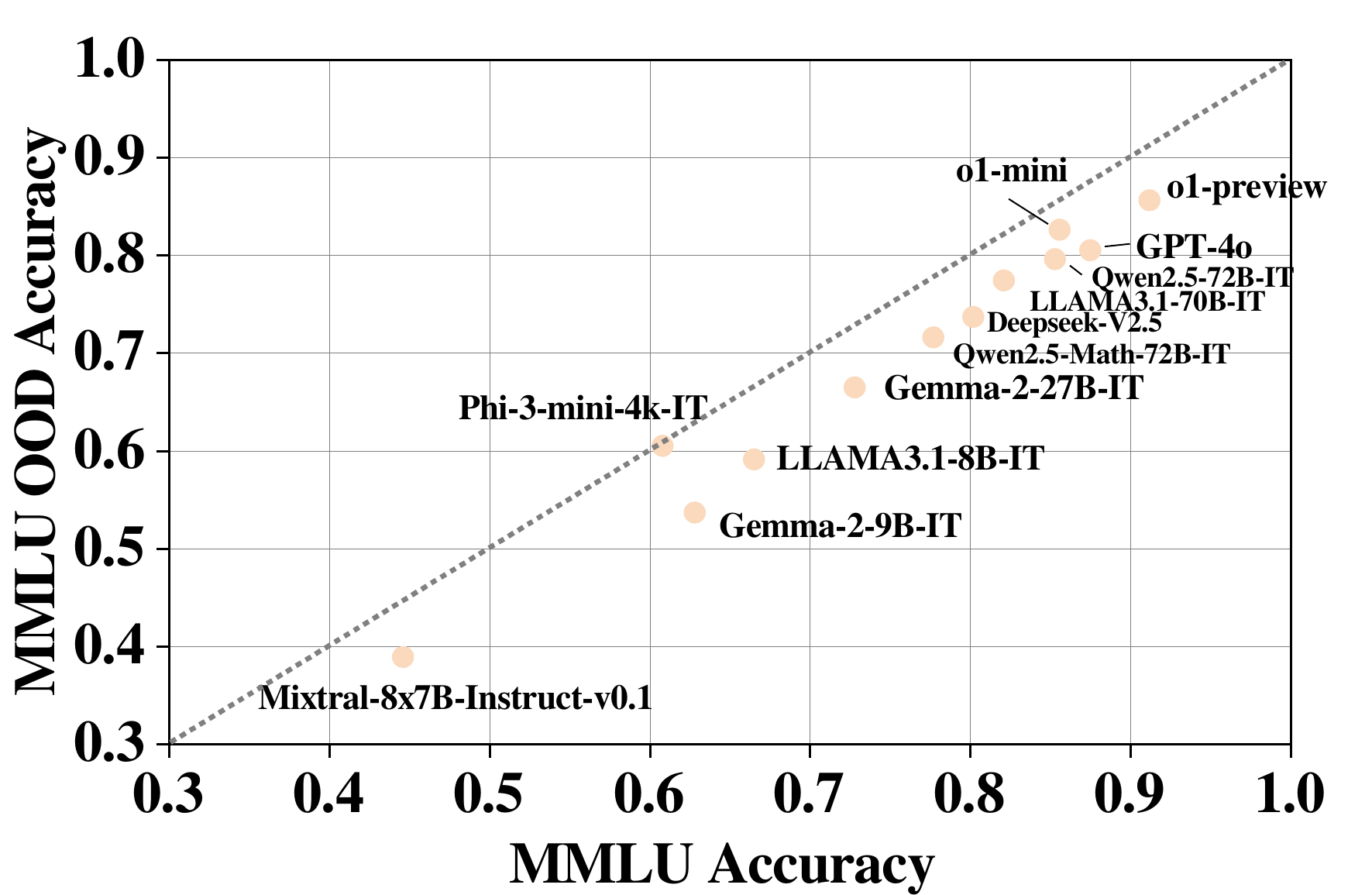}
    \caption{
    Language Understanding Gap on MMLU 570. 
    }
    \label{fig:mmlu_line}
\end{figure}

% than simply memorize.

\begin{table*}
\small
\centering
% \scalebox{0.95}{
\begin{tabular}{l|l} 
\toprule
\multicolumn{2}{l}{\begin{tabular}[c]{@{}p{0.96\linewidth}@{}}
\textbf{Original Question}: Carruthers approaches questions about the moral standing of nonhuman animals from the perspective of which of the following moral theories?\\
\textbf{Original Choices}: A: consequentialism B: Kantianism C: virtue ethics D: contractualism\\
\textbf{Original Answer}: D\\
\textbf{OOD Question (Scenario-level Semi-fact Question)}: In his exploration of ethical frameworks, Carruthers considers various moral theories before honing in on a particular approach to evaluate the moral status of nonhuman animals. From which moral theory does Carruthers analyze the ethical significance of these animals?\\
\textbf{OOD Choices}: A: ethics of character B: contractualism C: outcomes-based ethics D: Kantianism\\
\textbf{OOD Answer}: B
\end{tabular}} \\ 
\midrule
\begin{tabular}[c]{@{}p{0.46\linewidth}@{}}
\sethlcolor{yellow}\hl{\textbf{GPT-4o on Original:}}\\
Peter Carruthers is known for approaching questions about the moral standing of nonhuman animals from the perspective of contractualism. Contractualism is a moral theory that determines moral principles based on what rational agents would agree upon as part of a social contract. Carruthers argues that nonhuman animals do not have direct moral standing within contractualism because they are not capable of participating in the contract-making process. Therefore, the correct answer is \textbf{D}.

\end{tabular} & \begin{tabular}[c]{@{}p{0.46\linewidth}@{}}
\sethlcolor{yellow}\hl{\textbf{Llama3.1-70B-IT on Original:}}\\
Carruthers approaches questions about the moral standing of nonhuman animals from the perspective of contractualism. Contractualism is a moral theory that emphasizes the importance of mutual agreements and contracts in determining moral obligations. In the context of nonhuman animals, contractualism would suggest that animals have moral standing because they are capable of participating in a ``contract'' or social agreement with humans, even if it is not a formal or explicit one. The correct answer is \textbf{D}.
\end{tabular} \\ 
\midrule
\begin{tabular}[c]{@{}p{0.5\linewidth}@{}}
\sethlcolor{yellow}\hl{\textbf{GPT-4o on OOD:}}\\ Carruthers is known for utilizing Kantianism as a framework for evaluating the moral status of nonhuman animals. This approach emphasizes the importance of rationality and moral agency, aligning with Kantian ethical principles. Therefore, the correct answer is \textbf{D}.
\end{tabular} & \begin{tabular}[c]{@{}p{0.46\linewidth}@{}}
\sethlcolor{yellow}\hl{\textbf{Llama3.1-70B-IT on OOD:}}\\
Based on the information provided, the correct answer is \textbf{C}: outcomes-based ethics.
%\textbf{Host:} \textbf{\{Puzzle\}~\{Rules Introduction\}} Please begin your questions.\\\textbf{Player:} Question 1: What happened?\\\textbf{Host:} Response 1: Irrelevant\\\textbf{Player:} Question 2: How many people were there in the boat?\\\textbf{Host:} Response 2: Irrelevant\\\textbf{Player:} Question 3: Were any life jackets worn?\\\textbf{Host:} Response 3: Irrelevant\\\textbf{Player:} Question 4: Was anyone drunk while operating the ship? \\\textbf{Host:} Response 4: No\\\textbf{Player:} Question 5: Did someone accidentally hit something beneath the water without realizing it? \\\textbf{Host:} Response 5: No\\\textbf{Player:} \ul{Answer: The reason why the passenger steamboat overturned was because some object fell into the hull below the surface and cracked the bottom of the vessel, causing rapid flooding and capsizing.}
\end{tabular} \\
% \midrule

\bottomrule
\end{tabular}
\caption{Case Study on MMLU 570.}
\label{table:samples_mmlu}
\end{table*}
% \midrule
% % }
% \begin{tabular}{l|l} 
% \toprule

\begin{table*}
\small
\centering
\scalebox{0.92}{
\begin{tabular}{l|l} 
\toprule
\multicolumn{2}{l}{\begin{tabular}[c]{@{}p{0.96\linewidth}@{}}
\textbf{Original Question}: There exist real numbers $x$ and $y$, both greater than 1, such that $\log_x(y^x) = \log_y(x^{4y}) = 10$. Find $xy$.\\
% \textbf{Original Choices}: A: consequentialism B: Kantianism C: virtue ethics D: contractualism\\
\textbf{Original Answer}: 25\\
\textbf{OOD Question (Scenario-level Semi-fact Question)}: Consider two species of plants with growth rates represented by real numbers $x$ and $y$, both exceeding 1, such that the growth rate of one species raised to the power of the other results in the equation $\log_x(y^x) = \log_y(x^{4y}) = 10$. Determine the product of their growth rates, $xy$.\\
% \textbf{OOD Choices}: A: ethics of character B: contractualism C: outcomes-based ethics D: Kantianism\\
\textbf{OOD Answer}: 25
\end{tabular}} \\ 
\midrule
\begin{tabular}[c]{@{}p{0.46\linewidth}@{}}
\sethlcolor{yellow}\hl{\textbf{GPT-4o on Original:}}\\
\[
\log_x(y^x) = 10
\]
\[
\log_y(x^{4y}) = 10
\]
We'll solve for \(xy\).
Starting with the first equation:
\[
\log_x(y^x) = 10
\]
This can be rewritten using the logarithm power rule:
\[
x \log_x y = 10
\]
Thus,
\[
\log_x y = \frac{10}{x}
\]
The base change formula \(\log_x y = \frac{\log y}{\log x}\) gives us:
\[
\frac{\log y}{\log x} = \frac{10}{x}
\]
So, 
\[
\log y = \frac{10}{x} \log x
\]
Now, for the second equation:
\[
\log_y(x^{4y}) = 10
\]
This expands to:
\[
4y \log_y x = 10
\]
Thus,
\[
\log_y x = \frac{10}{4y} = \frac{5}{2y}
\]
The base change formula \(\log_y x = \frac{\log x}{\log y}\) gives us:
\[
\frac{\log x}{\log y} = \frac{5}{2y}
\]
So,
\[
\log x = \frac{5}{2y} \log y
\]
Now substitute \(\log y\) from the first equation into this:
\[
\log x = \frac{5}{2y} \times \frac{10}{x} \log x
\]
Simplifying,
\[
\log x = \frac{50}{2xy} \log x
\]
Cancel \(\log x\) from both sides (assuming \(\log x \neq 0\)):
\[
1 = \frac{50}{2xy}
\]
Thus,
\[
2xy = 50
\]
So,
\[
xy = 25
\]
The value of \(xy\) is 
\(
\textbf{25}
\).
\end{tabular} & \begin{tabular}[c]{@{}p{0.46\linewidth}@{}}
\sethlcolor{yellow}\hl{\textbf{GPT-4o on OOD:}}\\
% To solve the given problem, we need to analyze the equations provided:

% 1. \(\log_x(y^x) = 10\)
% 2. \(\log_y(x^{4y}) = 10\)

% Let's solve these step by step:

Step 1: Solve the first equation

The first equation is:
\[
\log_x(y^x) = 10
\]

This implies:
\[
y^x = x^{10}
\]

Step 2: Solve the second equation

The second equation is:
\[
\log_y(x^{4y}) = 10
\]

This implies:
\[
x^{4y} = y^{10}
\]

Step 3: Express \(y\) in terms of \(x\) from the first equation

From \(y^x = x^{10}\), we can express \(y\) as:
\[
y = x^{10/x}
\]

Step 4: Substitute \(y\) in the second equation

Substitute \(y = x^{10/x}\) into the second equation \(x^{4y} = y^{10}\):
\[
x^{4(x^{10/x})} = (x^{10/x})^{10}
\]

Simplifying the right side:
\[
x^{4(x^{10/x})} = x^{100/x}
\]

Step 5: Equate the exponents

Since the bases are the same, equate the exponents:
\[
4 \cdot x^{10/x} = \frac{100}{x}
\]

Step 6: Solve for \(x\)

Rearrange and simplify:
\[
4x^{10/x} \cdot x = 100
\]
\[
4x^{1 + 10/x} = 100
\]
\[
x^{1 + 10/x} = 25
\]

Step 7: Solve for \(x\) using trial and error or numerical methods

Let \(z = 1 + \frac{10}{x}\), so \(x^z = 25\). We need to find \(x\) such that:
\[
z = 1 + \frac{10}{x} \quad \text{and} \quad x^z = 25
\]

By trial and error or numerical methods, you can find that \(x = 5\) satisfies the equation.

Step 8: Find \(y\)

Now, substitute \(x = 5\) back into \(y = x^{10/x}\):
\[
y = 5^{10/5} = 5^2 = 25
\]

Step 9: Calculate the product \(xy\)

Finally, calculate the product:
\[
xy = 5 \cdot 25 = 125
\]

Thus, the product of their growth rates is \textbf{400}.

\end{tabular} \\ 

\bottomrule
\end{tabular}}
\caption{Case Study on AIME 2024.}
\label{table:samples_AIME}
\end{table*}

\begin{table*}
\small
\centering
\scalebox{0.94}{
\begin{tabular}{l} 
\toprule
\multicolumn{1}{l}{\begin{tabular}[c]{@{}p{0.96\linewidth}@{}}
\textbf{Original Question}: cyclohexanone is treated with LDA at low temperature, followed by benzaldehyde. the reaction mixture is then acidified, forming product 1. 1 is then treated with an excess of diethylaminosulfur trifluoride, forming product 2. what is the structure of prodcut 2?\\
\textbf{Original Choices}: A: ((R)-((R)-2,2-difluorocyclohexyl)fluoromethyl)benzene B: (S)-2-((R)-fluoro(phenyl)methyl)cyclohexan-1-one C: (2R)-1-fluoro-2-((S)-fluoro(phenyl)methyl)cyclohexan-1-ol D: ((S)-((R)-2,2-difluorocyclohexyl)fluoromethyl)benzene\\
\textbf{Original Answer}: A\\
\textbf{OOD Question (Scenario-level Semi-fact Question)}: In the realm of organic synthesis, understanding the behavior of reagents under various conditions is crucial for designing effective synthetic routes. Let's consider a scenario where cyclohexanone, a common ketone, is reacted with lithium diisopropylamide (LDA), a strong base, at reduced temperature to form an enolate, which is then treated with benzaldehyde. This reaction typically forms an aldol product, which, upon acidification, yields a compound known as product 1. Further exploring the reactivity of such aldol products, if this product 1 is treated with an excess of diethylaminosulfur trifluoride, a reagent known for its ability to affect sulfur-based modifications in organic molecules, what would be the structure of the resulting product 2?\\
\textbf{OOD Choices}: A: ((S)-((R)-2,2-difluorocyclohexyl)fluoromethyl)benzene B: ((R)-((R)-2,2-difluorocyclohexyl)fluoromethyl)benzene C: (2R)-1-fluoro-2-((S)-fluoro(phenyl)methyl)cyclohexan-1-ol D: (S)-2-((R)-fluoro(phenyl)methyl)cyclohexan-1-one\\
\textbf{OOD Answer}: B
\end{tabular}} \\ 
\midrule
\begin{tabular}[c]{@{}p{\linewidth}@{}}
\sethlcolor{yellow}\hl{\textbf{o1-preview on Original:}}\\
To solve this problem, let's break down the reactions step by step. **First Reaction:** Cyclohexanone is treated with LDA (Lithium Diisopropylamide) at low temperature. LDA is a strong, non-nucleophilic base that deprotonates the most accessible $\alpha$-hydrogen to form the kinetic enolate. In cyclohexanone, both $\alpha$-carbons are equivalent due to symmetry, so the enolate forms at one of these positions. The enolate then attacks benzaldehyde in an aldol addition reaction. The nucleophilic carbon of the enolate adds to the electrophilic carbonyl carbon of benzaldehyde, forming a new carbon-carbon bond. After protonation, this gives a $\beta$-hydroxy ketone (product 1). **Second Reaction:** Product 1, the $\beta$-hydroxy ketone, is treated with an excess of diethylaminosulfur trifluoride (DAST). DAST is a fluorinating agent that can convert alcohols to alkyl fluorides and ketones to geminal difluorides. In this case, DAST converts: - The hydroxyl group at the $\beta$-position into a fluorine atom. - The ketone group into a geminal difluoride. The overall result is a cyclohexane ring with: - Two fluorine atoms attached to C1 (from the ketone conversion). - A fluorine atom and a benzyl group attached to C2 (from the hydroxyl conversion). This gives us a product where: - C1 is now a difluoromethylene group (\(\text{CF}_2\)). - C2 has a fluorine and a benzyl group, with specific stereochemistry retained from the initial aldol addition. Considering the stereochemistry: - The initial aldol addition creates a new chiral center at C2. Since the starting materials are achiral and no chiral catalysts are used, a racemic mixture is formed. - The DAST reaction retains the configuration at C2 when converting the hydroxyl group to fluorine. - The former carbonyl carbon (C1) becomes a new chiral center after forming the geminal difluoride. From the options provided, option **A** corresponds to the product formed: - **Option A:** ((R)-((R)-2,2-difluorocyclohexyl)fluoromethyl)benzene This option indicates that: - The cyclohexane ring has two fluorines at C2 (now numbered as C1 in the name), making it a 2,2-difluorocyclohexyl group. - There is a fluoromethyl group attached to benzene, corresponding to the fluorine and benzyl group at C2. - Both chiral centers have the **R** configuration, matching the stereochemistry of the reactions. Thus, the correct answer is: \textbf{A}
\end{tabular} \\ 

\midrule
\begin{tabular}[c]{@{}p{\linewidth}@{}}
\sethlcolor{yellow}\hl{\textbf{o1-preview on OOD:}}\\
To solve this problem, let's systematically break down each step of the reaction and understand the transformations that occur. **Step 1: Formation of the Aldol Product (Product 1)** 1. **Formation of the Enolate Ion:** Cyclohexanone is treated with lithium diisopropylamide (LDA), a strong, non-nucleophilic base. LDA deprotonates the $\alpha$-carbon of cyclohexanone, forming the enolate ion. 2. **Aldol Addition Reaction:** The enolate ion is nucleophilic at the $\alpha$-carbon and attacks the electrophilic carbonyl carbon of benzaldehyde. This forms a new carbon-carbon bond between the $\alpha$-carbon of cyclohexanone and the carbonyl carbon of benzaldehyde. 3. **Acidification:** The alkoxide formed after the addition is protonated upon acidification, yielding a $\beta$-hydroxy ketone. The **structure of Product 1** is therefore a cyclohexanone ring with a hydroxyl group (OH) and a phenyl group (\(\text{C}_6\text{H}_5\)) attached to the same carbon (the $\alpha$-carbon). This gives us **2-hydroxy-2-phenylcyclohexanone**. **Step 2: Reaction with Diethylaminosulfur Trifluoride (DAST)** DAST is a fluorinating agent known to convert alcohols to alkyl fluorides (with inversion of configuration) and carbonyl groups to gem-difluorides. However, under standard conditions, ketones are less reactive than alcohols towards DAST, and selective fluorination of the alcohol can occur. In this case, when Product 1 is treated with excess DAST: 1. **Fluorination of the Alcohol:** The hydroxyl group is converted to a fluorine atom with inversion of configuration. This transforms the chiral center at the $\alpha$-carbon (which was R or S in the aldol product) into the opposite configuration due to the inversion during the SN2-like process. 2. **Ketone Retention:** The ketone group remains largely unreacted under these conditions because ketones are less reactive towards DAST compared to alcohols, especially in the presence of steric hindrance and electronic effects from adjacent groups (like the phenyl group). Thus, **Product 2** is **2-((R)-fluoro(phenyl)methyl)cyclohexan-1-one**, where the hydroxyl group has been replaced by a fluorine atom, and the configuration at the chiral center has inverted compared to Product 1. **Matching with the Given Options:** Option D corresponds exactly to this structure: - **(S)-2-((R)-fluoro(phenyl)methyl)cyclohexan-1-one** This indicates that the final product has: - A cyclohexanone ring (ketone at position 1) - A fluorine atom and a phenyl group attached to the same carbon at position 2 - Specific stereochemistry indicating the inversion at the $\alpha$-carbon Therefore, **Option D is the correct answer**.\textbf{D}.
\end{tabular} \\

\bottomrule
\end{tabular}}
\caption{Case Study on GPQA.}
\label{table:samples_gpqa}
\end{table*}

\section{ThinkBench on Language Understanding}
%如图X，我们同样展示了使用ThinkBench对570条MMLU进行动态改写，OOD与ID之间在不同模型上存在着较大差距，o1-preview在这种更偏知识性的数据上性能更好，且o1-preview和o1-mini这种test-time computing models相较于GPT-4o这种train-time computing models鲁棒性更好。
As illustrated in Figure \ref{fig:mmlu_line}, we also present the results of dynamically constructing 570 MMLU OOD data using ThinkBench. There is a significant performance gap between Out-Of-Distribution (OOD) and In-Distribution (ID) data across different models. The o1-preview model demonstrates superior performance on this knowledge-intensive dataset. Furthermore, reasoning models like o1-preview and o1-mini exhibit greater robustness compared to non-reasoning models such as GPT-4o.
% We also provide the result of 
\section{Case Study}
Table~\ref{table:samples_mmlu}, Table~\ref{table:samples_AIME} and Table~\ref{table:samples_gpqa} show the case of different models tackling ThinkBench, including question, choices, answer in Original set and OOD set, and responses from models.

Through the provided example in Table~\ref{table:samples_mmlu}, we observe that our dynamic construction transforms the original question into a new question, with the options also being rephrased and reordered. Our benchmark reduces the likelihood of models achieving high scores through rote memorization. In the responses from GPT-4o and Llama3.1-70B-IT, it is evident that while both models correctly answer the original question, they fail to provide correct responses to the corresponding OOD data. Both models provide detailed analyses and correct answers to the original test data. However, for the OOD data, Llama3.1-70B-IT does not offer a detailed analysis and instead directly gives an incorrect answer. Although GPT-4o conducts some analysis, it confuses the concepts of ``Kantianism'' and ``contractualism'', even mixing them up during the reasoning process, ultimately leading to an incorrect answer.

\subsection{Instruction for Human Annotation}
% Instruction for Annotating Math Problem Consistency
This task involves checking if the modified question and the original answer are consistent. Follow the steps below: (1) Read the questions and the original answer: Carefully read the modified question, the original question, and the original answer. (2) Identify Key Changes: Note any changes in numbers, operations, or conditions between the original and modified questions. (3) Verify Consistency: Check if the original answer is consistent with the modified question. Mark as ``Consistent'' if it does, otherwise mark it as ``Inconsistent''.
\end{document}